\documentclass[journal]{IEEEtran}
\usepackage{graphicx}
\usepackage{subcaption}
\usepackage{booktabs} 
\usepackage{amssymb}
\usepackage{amsmath}
\usepackage{lineno}

\usepackage{algpseudocode,algorithm,algorithmicx}
\usepackage{setspace}
\usepackage{siunitx}
\usepackage[dvipsnames,table]{xcolor}
\usepackage{caption}
\usepackage{epsfig} 
\usepackage{array}
\usepackage{multirow}

\definecolor{LightCyan}{rgb}{0.78,1,1}
\title{\fontsize{20}{20}\selectfont{A Method for Handling Multi-class
Imbalanced Data by Geometry based Information Sampling and Class
Prioritized Synthetic Data Generation (GICaPS)}}
\author{Anima Majumder$^1$ \thanks{$^{1}$ (anima.majumder, d.samrat, swagat.kumar)@tcs.com, Tata Consultancy Services, Bangalore, India,}, Samrat Dutta$^1$, Swagat Kumar$^1$, Laxmidhar Behera$^2$
\thanks{$^{2}$ lbehera@iitk.ac.in, IIT Kanpur, U.P., India.} }

\begin{document}
\maketitle
\begin{abstract}
This paper looks into the problem of handling imbalanced data in a
multi-label classification problem. The problem is solved by proposing
two novel methods that primarily exploit the geometric relationship
between the feature vectors. The first one is an undersampling
algorithm that uses angle between feature vectors to
select more informative samples while rejecting the less informative
ones. A suitable criterion is proposed to define the informativeness
of a given sample. The second one is an oversampling algorithm that
uses a generative algorithm to create new synthetic data that respects
all class boundaries. This is achieved by finding \emph{no man's land}
based on Euclidean distance between the feature vectors. The efficacy
of the proposed methods is analyzed by solving a generic multi-class
recognition problem based on mixture of Gaussians.  The superiority of
the proposed algorithms is established through comparison with other
state-of-the-art methods, including SMOTE and ADASYN,  over ten
different publicly available datasets exhibiting high-to-extreme data
imbalance. These two methods are combined into a single data
processing framework and is labeled as ``GICaPS'' to highlight the role of
geometry-based information (GI) sampling and Class-Prioritized
Synthesis (CaPS) in dealing with multi-class data imbalance problem,
thereby making a novel contribution in this field. 
\end{abstract}
\IEEEpeerreviewmaketitle
\section{Introduction} \label{sec:intro}

\IEEEPARstart{M}{ajority} of existing classification techniques,
including deep learning approaches are typically designed to perform
well when the distribution of data among classes is balanced.
\textcolor{black}{Many of these methods perform poorly on real-world
  datasets that are inherently \emph{class-imbalanced}
  \cite{leevy2018survey} \cite{ali2015classification}
  \cite{rout2018handling} \cite{ali2019imbalance} with the
  \emph{majority} class(es) forming the bulk of the dataset while a
  disproportionately smaller share coming from the \emph{minority}
  class(es). Spam filtering, network intrusion detection, cancer
  diagnosis, detecting fraudulent transaction are some of the
  applications that generate imbalanced datasets. Classifiers trained
  on such imbalanced datasets are biased towards the majority class
  making them unreliable for use in several cases where the minority
  class is of critical interest. For instance, it is very important to
  detect a fraudulent transaction even when its occurrence is rare
  compared to the overall number of genuine transactions made over a
  given period.  Same applies to the case of medical diagnosis where a
  single case of false negative (e.g., failing to detect a malignant
  tumor) can lead to serious consequences even when the overall
  classifier accuracy is more than 99\%. Many of these applications
  exhibit high-to-extreme class imbalance where the
  majority-to-minority class ratio could be more than 10,000:1
  thereby, posing serious learning challenges for classifier design
  \cite{krawczyk2016learning} \cite{leevy2018survey}. } 

\textcolor{Black}{Most of the existing methods for dealing with
  imbalanced data can be broadly classified into three categories:
  \emph{data-level methods}, \emph{algorithm-level methods} and
  \emph{hybrid} approaches \cite{ali2015classification}
  \cite{krawczyk2016learning} \cite{leevy2018survey}.
  \emph{Data-level methods} focus on improving the dataset by using
  methods such as over- or under-sampling, feature selection etc. Some
  of the popular data-sampling approaches include Random Over-Sampling
  (ROS), Random Under-Sampling (RUS), SMOTE (Synthetic Minority
  Over-Sampling Technique) and its variants \cite{chawla2002smote}
  \cite{han2005borderline} \cite{chawla2003smoteboost} and, ADASYN
  \cite{he2008adasyn}. On the other hand, \emph{algorithm-level
  methods} try to learn the imbalance data distribution from the
  classes in the datasets. These methods can be further sub-grouped
  into \emph{cost-sensitive methods} and \emph{ensemble methods}. The
  former works by assigning varying cost or weight to different
  instances or classifiers in the event of misclassification while the
  later combines the output of multiple classifiers built on the
  dichotomies created from the original  dataset to improve the
  classification performance. Bagging and Boosting
  \cite{wang2016online} \cite{vezhnevets2005modest} are two common
  types of ensemble methods. The \emph{hybrid methods} combine the
  advantage of data-level and algorithm-level methods and usually
  employ more than one machine learning algorithms to improve the
  classification accuracy \cite{lee2010hybrid}
  \cite{wozniak2013hybrid}\cite{lee2007features}
  \cite{tahir2012inverse} \cite{zhang2018approach}.  Many of these
  algorithms have been developed to solve the binary-class data
  imbalance problem which has been studied more extensively compared
  to the multi-class imbalance problem \cite{krawczyk2016learning}. A
  multi-class data imbalance problem is usually solved by using some
  decomposition strategy to reduce it to a set of binary-class problems
  which can be solved by using one of the above techniques
  \cite{zhang2019multi} \cite{krawczyk2016learning}
  \cite{leevy2018survey}}. 

\textcolor{Black}{In this paper, we restrict our discussion to
  sampling-based methods that form a major part of the data-level
  methods for imbalanced data learning. The data-sampling methods
  could be further divided into sub-groups, namely, over-sampling and
  under-sampling methods. While the former aims at adding or
  replicating instances of the minority class, the later focusses on
  removing instances of the majority class in a given dataset to
  reduce to overall data imbalance. The replication or removal of data
  is either done randomly (e.g. ROS /RUS \cite{charte2015addressing}
  \cite{seiffert2009rusboost} \cite{tahir2012inverse}) or through an
  intelligent algorithm (e.g.  SMOTE \cite{chawla2002smote}, ADASYN
  \cite{he2008adasyn}). The data-sampling algorithms are designed to
  address problems such as overfitting (during oversamping), loss
  of valuable information (during undersampling) or the existence of
  disjuncts (imbalance within a class). The problem becomes more
  challenging when the dataset exhibits considerable class overlap
  \cite{ali2015classification} \cite{lee2018overlap}. Overlapping
  classes have low degree of separability between the classes and the
  data points on the boundaries may belong to any of the overlapping
  classes. The class overlapping becomes a more serious issue in the
  presence of sensor noise or outliers \cite{gupta2018handling}.
  Various methods have been proposed to address the class overlapping
  issue within the imbalanced data classification problem. These methods
  span across the categories mentioned above with varying degree of
  success. For instance, the authors in \cite{saez2019addressing} suggest
a one-vs-one decomposition strategy to alleviate the presence of
overlapping without modifying existing algorithms for modifying the
dataset. Similarly, the authors in \cite{lee2018overlap} propose a hybrid
approach that combines fuzzy SVM with a k-nearest algorithm to deal
with data imbalance and class overlapping simultaneously. Several
variants of SMOTE \cite{fernandez2018smote} have been proposed to
address the class separability issue in imbalanced learning problems.
In another work \cite{yang2019manifold}, authors use manifold distance
instead of Euclidean distance to improve the class separability of the
samples. In spite of these efforts, the imbalanced learning is still
considered as a challenging problem particularly when there exists
significant class overlapping. } 

\textcolor{Black}{In this paper, we aim to address the problem of
  class separability for a multi-class imbalanced learning problem.
  This is achieved by using a combination of oversampling and
  undersampling method that use geometric information sampling and
  class prioritized synthesis that not only improves the separability
  between classes but also increases the diversity of samples within
  each class by selectively removing redundant samples. The proposed
  method is, hence, termed GICaPS which is an acronym for Geometric
  Information-based Sampling and Class-Prioritized Synthesis. The
  undersampling approach uses distance in the polar (angular)
  coordinates to remove data points from a majority while ensuring
  that valuable information is not lost by constraining the removal of
  samples only from other orthants (n-dimensional orthogonal
  half-spaces). Similarly, the oversampling algorithm creates
  synthetic data samples in a minority class that respects class
  boundaries. This is ensured by avoiding data generation in regions
  with high class interference, otherwise known as `\emph{no man's
  land}'. This helps in addressing the class overlapping problem which is
  particularly challenging when the data has noise and outliers.  The
  above oversampling and undersampling approaches can be used
  separately or together for a given problem. The proposed data
  processing framework, called  GICaPS, is shown to provide superior
  classification results compared to many of the existing
  state-of-the-art methods over 10 different publicly available
datasets having very high class imbalance. }

  \textcolor{Black}{In short, the major contributions made in this paper are as follows: 
    \begin{itemize}
      \item A novel undersampling approach that uses angular distance
        to remove samples in a majority class while constraining the
        removal of points only from other orthants. The use of angular
        information for undersampling has not been explored before and
        hence forms a novel contribution in this paper. 
      \item A novel oversampling algorithm is provided that respects
        class boundaries and addresses the class overlapping problem
        by avoiding data synthesis in the regions with high class
        interference, otherwise known as \emph{no man's land}.
        Mathematical formulation for identifying no-man's land is
        provided and to the best of our knowledge, such a concept has
        not been used before in this context. 
      \item The efficacy of the above approaches is established
        through rigorous analysis on ten different publicly available
        datasets exhibiting very high class imbalance and is shown to
        outperform many of the existing state-of-the-art methods in
        this field. 
    \end{itemize}}

The rest of the paper is organized as follows. An overview of related
work is provided in the next section. The datasets used in
this paper are discussed in Section \ref{sec:dataset_overview}. The
proposed undersampling and oversampling approaches are explained in
Section \ref{sec:prop_app}. The working of the proposed methods is
demonstrated on a simulated 3-dimensional data using a Gaussian
Mixture Classifier and is described in Section \ref{subsec:pain_gmm}.
The efficacy of the proposed method is further established by
providing comparison with other state-of-the-art methods on ten
different datasets that exhibiting high level of imbalance. This along
with other analyses are discussed in Section \ref{sec:exp_res}.
Finally, the conclusion and future scope of this work is discussion in
Section \ref{sec:concl}.    

\section{Related work}
Handling of imbalance data is a long-lasting problem and lots of work have been done in this area since last few decades \cite{xie2020gaussian}. Based on the literature, the approaches of solving this problem can be broadly categorized into three different groups:data-level, cost-sensitive and ensemble learning approaches. Data-level approaches try to balance skewed distribution by using various resampling approaches. Again, it can be categorized into two sub-groups: undersampling and oversampling. Cost-sensitive learning \cite{tang2008svms, zhang2016transfer} prioritizes accurate classification of minority class samples over majority class/es. Target of this approach is to generate a classification model with minimum cost and this is achieved by establishing a cost matrix in which the elements of the matrix indicate the penalty strength for the instances that are misclassified \cite{xie2020gaussian}. Ensemble learning approaches \cite{liu2008exploratory, zhu2018geometric, zhu2020globalized} combine multiple base classifiers in order to achieve promising results on imbalanced datasets. Various hybrid models are also there in the literature that combines data-level techniques with ensemble learning, resulting into improved performance in imbalanced data classification\cite{chawla2003smoteboost, lim2016evolutionary}. Our approach of handling imbalanced data falls under the category of data-level approaches. Rest of this section will thus concentrate on data-level based undersampling and oversampling approaches. 

\label{sec:related_work}
\subsection{Previous works on undersampling}
A majority class in an imbalanced dataset often contains a lot of redundant or less informative data which increases unnecessary computation cost and also mis-leads the classifier in predicting accuracy. 
However, unlike, oversampling approaches, not much works have been done in the direction of undersampling. 
One of the earliest undersampling technique used to alleviate the problem of class imbalance in the dataset is Random Under Sampling (RUS) \cite{yen2009cluster}.  However, one  major drawback of this approach is that it may potentially discard useful information while sampling the data. In real-life datasets, the distribution of data can be such that, it may contain data densely populated in some region and can have very sparse distribution in some other region. As RUS has almost equal probability of picking up a sample from anywhere within the distribution, the densely populated regions will remain dense even after undersampling, and the regions containing very sparse data distribution may lose very informative data.
Some research works are also performed in the direction of minimizing the effect of information loss occurs after RUS. EasyEnsemble \cite{liu2009easyensemble} and BalanceCascade \cite{raghuwanshi2019classifying} are two of such approaches. Another undersampling approach, commonly known as Edited Nearest Neighbour (ENN), was adopted from the study of Wilson \cite{wilson1972asymptotic}. ENN mainly focuses on instances near the decision boundary and selectively removes majority class instances by considering its k nearest neighbours that belong to the other class. Few extensions of this work include Neighbourhood Cleaning Rule (NCL) \cite{laurikkala2001improving,jo2004class, koziarski2017ccr}. Another data cleaning strategy was introduced by Kubat and Matwin that uses Tomek links \cite{kubat1997addressing} to remove only borderline majority samples.  All these aforesaid methods mainly focuses on the removal of borderline majority class data and overlooks the desire of removing unnecessary less informative and redundant data which may lie within the majority class. An attempt to remove the redundant data from the distribution of majority class set is presented in an approach commonly known as \emph{cluster centroid undersampling} \cite{rahman2013cluster}. Majority class is undersampled by forming clusters and the sampled data are chosen as cluster centroids. The number of clusters is set by the level of undersampling. Few extensions of this work include \cite{doi:10.1080/0740817X.2015.1110269, takeshita2017discriminative}. Zhao er al. \cite{doi:10.1080/0740817X.2015.1110269} applied an unsupervised learning algorithm that transforms the classification problem into several classification sub-problems. K-medoids based undersampling approach is applied in \cite{takeshita2017discriminative} and only the cluster centers are considered as sampled data. Again, all these approaches can't ensure retaining of most informative data as they replaces the real-samples with the cluster centers. In contrast, we propose an undersampling technique based on the angular information among the feature vectors of majority class, that ensures retaining of more informative data and removal of less informative or redundant data from all the regions within the distribution set. 

\subsection{Previous works on oversampling}
Although data balancing can be best handled by the implication of both oversampling and undersampling approaches, however, researchers have been more frequently applying oversampling approaches to solve this problem. Lots of works have been done in data driven oversampling techniques \cite{wei2020ni, mani2003knn, chawla2002smote, han2005borderline, he2008adasyn, chen2010ramoboost, sharma2018synthetic}. Random Over-Sampling (ROS) with replacement \cite{mani2003knn} is the fundamental concept of oversampling techniques. This method selects a set of $E$ sampled minority class data from the minority set $S_{min}$ and then replicates those selected data into the set to make a balanced dataset. As the replacement process of ROS is completely random, this approach does not specify a clear borderline between any two classes. 
Moreover, simple replication of existing data into the original minority class/es can cause the problem of overfitting \cite{chawla2002smote}. Among the existing data-driven oversampling approaches, Sampling with Data Generation (SMOTE) \cite{chawla2002smote} is still considered as the \emph{state of the art} in the literature due to its simplicity and easy implementability. However, it is associated with various shortcomings. One important drawback of SMOTE is that, it does not consider possible interferences of data from other classes while generating synthetic examples, thereby increases the chances of multi-class overlapping and also introduces additional noises. Another major issue with the SMOTE is that, it has no control over the number of new data to be generated (it merely replicates same number of data, originally present in a minority class). Therefore, the dataset remains imbalanced even after applying SMOTE.  
Over the period of time various improvements have been done on SMOTE \cite{fernandez2018smote}. Some of those, include Borderline-SMOTE \cite{han2005borderline}, Adaptive Synthetic Sampling Approach for Imbalanced Learning (ADASYN) \cite{he2008adasyn}, Ranked minority oversampling in boosting (Ramoboost) \cite{chen2010ramoboost}. ADASYN is developed based on the idea of Borderline-SMOTE. Unlike SMOTE, Borderline-SMOTE only creates synthetic samples for the data points which are near to the border, while taking into consideration that no synthetic data should be generated for "Noise" instances. ADASYN adapts the concept of Borderline-SMOTE and creates different number of synthetic data based on the data distribution. Unlike, SMOTE and Borderline-SMOTE, ADASYN algorithm can decide the number of synthetic examples that need to be generated for each minority examples by the number of its majority nearest neighbor, I.e., the more the majority nearest neighbor, the more synthetic examples will be created. One important drawback with this approach is that, the synthetic data is generated only near to the boundary. Secondly, It does not consider the possibility of interference of other minority or majority class data while generating synthetic data. Third, both SMOTE and ADASYN do not consider within class imbalance while generating new data. I.e, the data-intensive minority regions may still remain dense, while the data-sparse minority regions may remain sparse \cite{douzas2018improving}. In contrast, the proposed approach of oversampling mainly focuses on the inter-class borderline data over-lapping issues and effectively localizes all the possible data over-lapping regions between any two classes and wisely avoids those regions while synthetically generating new data within the targeted minority class. In addition to that, the proposed oversampling techniques reduces within class imbalance by using K-means clustering before synthetic data generation. Some research works on oversampling techniques that attempted to reduce within class imbalance are Cluster-SMOTE \cite{cieslak2006combating},MWMOTE (majority weighted minority oversampling technique) \cite{barua2012mwmote}, DBSMOTE (density-based synthetic minority oversampling
technique) \cite{bunkhumpornpat2012dbsmote}, CURE-SMOTE \cite{ma2017cure}, and K-means SMOTE \cite{douzas2018improving}. However, all these methods do not consider interference of other classes boundaries while interpolating data within the selected minority class. 
\section{Imbalanced data distribution in different datasets}
\label{sec:dataset_overview}
The proposed technique has the capability of handling highly imbalanced datasets with multiple classes.
Multi-class datasets used in our experiments are, Abalone, Glass, Wine, Shuttle and Pain with classes 23, 5, 3, 7 and 15 respectively. Remaining datasets contain binary class data. Table \ref{tab:diff_classs_data_dist} summarizes the details of ten different datasets used in our experiments.
\begin{table*}
\captionsetup{width=0.95\linewidth}
\centering
\footnotesize{
\begin{tabular}{SSSSSSS} \toprule
{Dataset} & {$\#$ } & {$\#$}& {$\#$ } & {Minority} & {Majority } \\ 
    {} & {Classes} & { Feature} & { Data} & {instances} & {instances}   \\ 
    
    \midrule
    \rowcolor{LightCyan}{Abalone}  & 23 &8 &4177 & 2 & 689 \\ \midrule
      {Spambase}  & 2 &58 &4601 & 1813 & 2788 \\ \midrule
       {Glass}  & 5 &10 &214 & 9 & 163 \\ \midrule
        {Ionosphere}  & 2 &34 &351 & 126 & 225 \\ \midrule
         {Sonar}  & 2 &60 &208 & 97 & 111 \\ \midrule
         {Wine}  & 3 &13 &178 & 48 & 71 \\ \midrule
          {Pima Indian diabetes}  & 2 &8 &768 & 268 & 500 \\ \midrule
          {Shuttle}  & 7 &9 &43500 & 6 & 34108 \\ \midrule
           {fertility}  & 2 &9 &100 & 12 & 88 \\ \midrule  
          \rowcolor{LightCyan} {Pain}  & 15 &22 &48198 & 5 & 39835 \\ \bottomrule            
  
\end{tabular}}
\caption{Imbalance nature of different datasets (highest number of data in majority class and least number of data minority class are given in the columns \emph{Majority instances} and \emph{Minority instances} respectively.) Among the given datasets, Abalone and Pain have highly imbalanced data with more number of classes ( $23$ and $15$ respectively).} 
\label{tab:diff_classs_data_dist}
\end{table*}
\begin{center}
 \begin{table}
\captionsetup{width=0.85\linewidth}
\caption{Data distribution of pain db. (The terms are: I-Intensity, DS-Data size)} 
\resizebox{\columnwidth}{!}{%
\centering\begin{tabular}{c c c c c c c c c} 
\hline\hline 
I & $I_0$ & $I_1$ & $I_2$ & $I_3$ & $I_4$ & $I_5$ & $I_6$ & $I_7$  \\  
 DS & 39835 & 2908 & 2349 & 1409 & 802 & 242 & 270 & 53 \\
 \hline 
 I & $I_8$ & $I_9$ & $I_{10}$ & $I_{11}$ & $I_{12}$ & $I_{13}$ & $I_{14}$ & $I_{15}$\\
  DS & 79 & 32 & 67 & 76 & 48 & 22 & 1 & 5\\
\hline 
\label{tab:pain_data_dist}
\end{tabular}
}
\end{table}
\end{center}
To demonstrate the amount of variations in data distribution among different classes, we present an example in the Table \ref{tab:pain_data_dist}. The Table \ref{tab:pain_data_dist} shows pain intensity (class) in the first row and corresponding sample size is given in the second row. Such highly skewed distributions of data among classes motivated us to apply both undersampling and oversampling technique to generate a balanced dataset. It is to be notated that, for synthetic generation of new data, an initial data distribution within a class must have at least two data. For instance, the class labeled as $14$, in the Pain database \cite{lucey2011painful} has only one data. We have thus, removed that label in our experiment. Following section presents detailed description of the proposed undersampling and oversampling technique. 
 
\section{Proposed Imbalanced data handling technique- GICaPS}
\label{sec:prop_app}
The earlier section presented a few instances of extremely imbalanced datasets where the majority classes dominate more than $90$ percent of the training data. Therefore, an attempt to balancing the dataset by merely generating synthetic data in the minority classes would make the dataset huge and thereby severely impact the computation during training. We observe that in extremely imbalanced datasets, the majority classes contain many repeated/ less informative occurrences which are in fact redundant to training the classifier. Hence, GICaPS is the combination of two algorithms: \textit{GICaPS-undersampling} and  \textit{GICaPS-oversampling}. 
\subsection{GICaPS undersampling approach}
The intention behind applying an undersampling approach to a majority class is to get rid of the data samples those have very less impact on the training of a network. This not only helps balancing the training data, also results in significant decrease in computational burden during network training. 
It is important to note that the similarity between two feature vectors is directly proportional to their dot product. Hence, by measuring the angle between two feature vectors, one can determine the dissimilarity or distinctiveness between them, which is in fact directly proportional to the angle between the vectors. The main focus of undersampling technique is to remove feature vectors from the majority class while maintaining the intra-class data diversity (as much as possible). However, maintaining diversity is difficult in random undersampling or euclidean distance based approach. Hence, we embrace a sorting technique in the angular space, where we ensure that each feature vector is distinct than the others. 
The proposed undersampling algorithm works using two principles:
\begin{itemize}
\item discard a feature vector if a similar feature vector already exists in the class
\item reduce the density of data samples in densely populated regions of the class while ensuring the 
left over features vectors are away from each other in the angular space by certain distance 
\end{itemize} 

The undersampling technique has the following attributes:
\begin{enumerate}
 \item The overlapped and less informative feature vectors are removed. 
 \item The feature vectors are picked up based on angular information between the vectors in the entire majority class. 
 \item The approach makes sure that a uniform intra-class data density is maintained within the new dataset of the majority class. 
 \item The approach is unique in a sense that it also considers the intra-plane information of the vectors while performing undersampling operation. To the best of our knowledge this has not been introduced anywhere else in the  literature.   
\end{enumerate}

\subsubsection{GICaPS undersampling technique}
This section gives a detailed explanation of the proposed undersampling approach.
\begin{figure}
\centering
\captionsetup{width=0.85\linewidth}
\epsfig{file=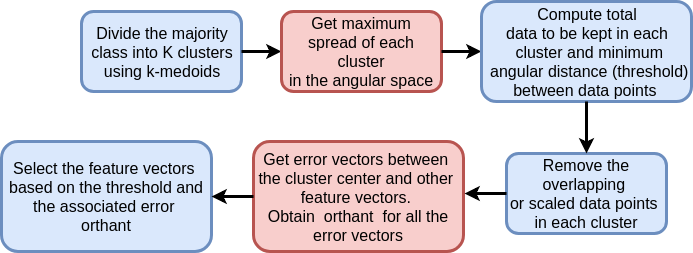,width=0.9\linewidth}
\caption{An overview of the proposed undersampling approach.}
\label{fig:block_undersam}
\end{figure}
 A block diagram presenting different steps involved in the proposed undersampling approach is shown in Figure \ref{fig:block_undersam}. Given a majority class, the entire class is divided into $K$ clusters with cluster centers $\mathbf{c}^k \in (1,2,...K)$. We use K-medoids to find the clusters in the majority class. The reason for using K-medoids is that it selects the cluster centers as an original data point while minimizing $\mathcal{L}_1$ norm over the entire data which makes it more robust than K-means to outliers. The choice of the number of clusters in the majority class is also important especially when the majority class has a heterogeneous data distribution. The optimal number of clusters are chosen using elbow rule \cite{elbo}, however one can use more efficient techniques as in \cite{silh} to determine the optimal number of clusters. The undersampling operation is performed in each of the $k$ clusters of the class. First the angle between the cluster center $\mathbf{c}^k$ and all other feature vectors ($\mathbf{x}_i$, $i=1,2,...$) is calculated and also the angle between every two features are calculated. Then the algorithm checks for the overlapped or similar feature vector and densely populated regions. The rejection criteria for a data sample are stated in the following.

\emph{Rejection criteria 1:} 
The first rejection criteria is set such that the feature vectors are separated at least by the minimum threshold angle in the angular space. Hence, we start by arranging the angles in ascending order, which places $\mathbf{c}^k$ at the top of the stack.  Then look for the first feature vector $\mathbf{x}_i$ from the remaining data in the stack which has angular displacement with the center that exceeds certain threshold. If the angle between a feature vector $\mathbf{x}_i$ and the cluster center $\mathbf{c}^k$ is greater than the threshold $\delta \sigma_k$ then keep the data else remove the feature vector. Keep removing the feature vectors until the initial threshold criteria is satisfied. 

\begin{figure}
\centering
\begin{subfigure}[b]{0.25\textwidth}
\includegraphics[scale=.1]{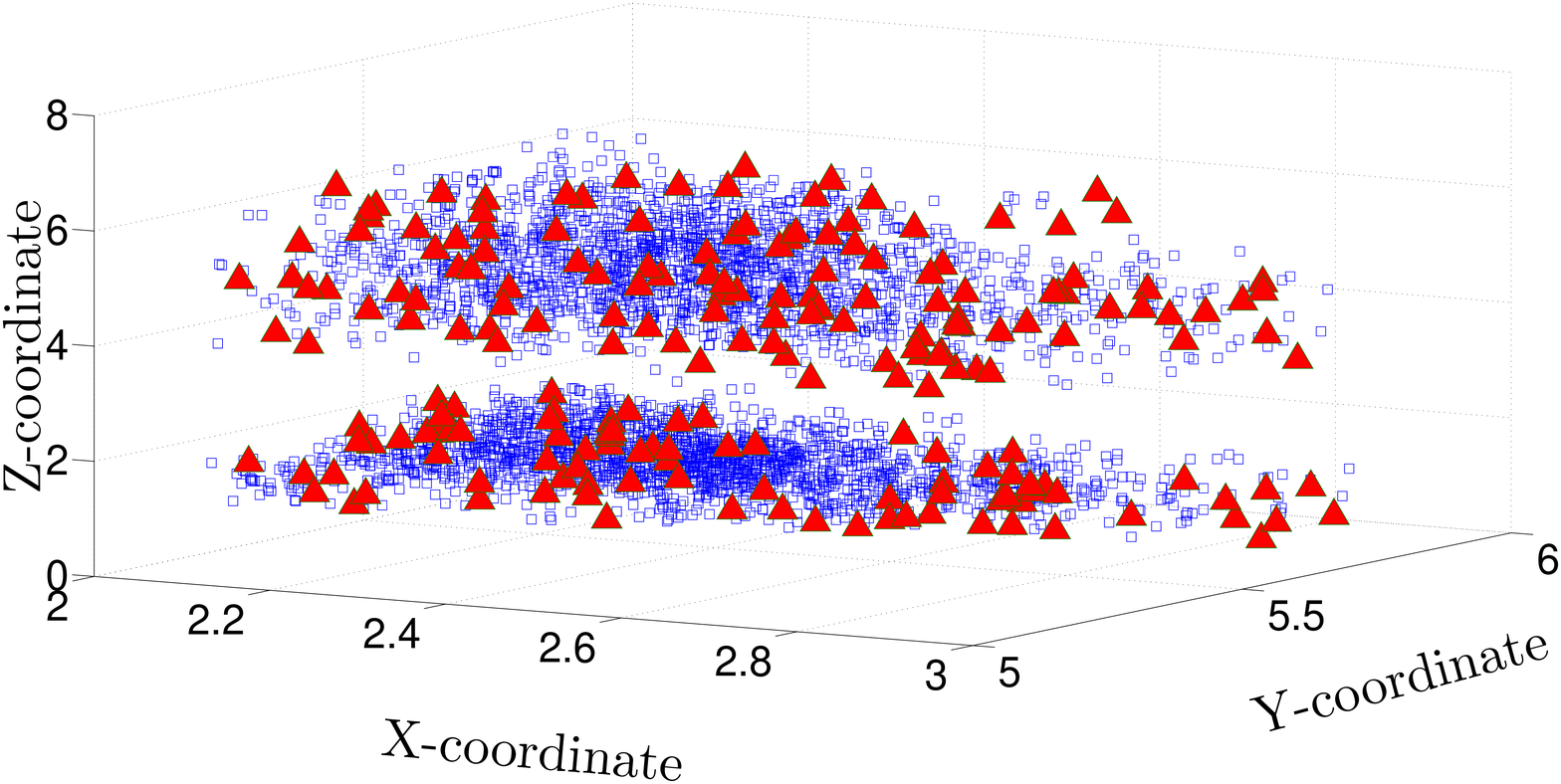}
\caption{}
\label{fig:under_sam1}
\end{subfigure}~
\begin{subfigure}[b]{0.25\textwidth}
\centering
\includegraphics[scale=.2]{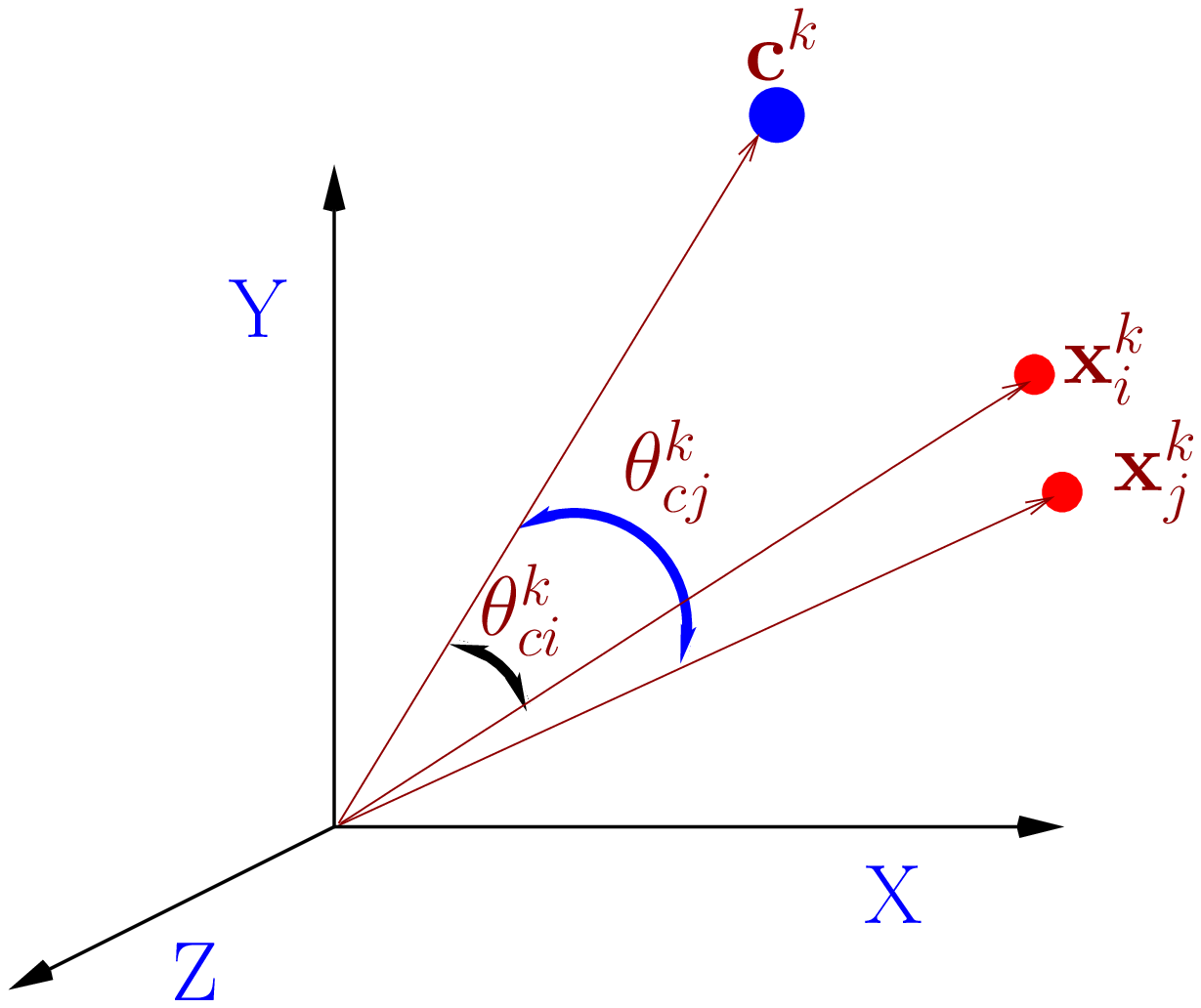}
\caption{}
\label{fig:3d_data_us}
\end{subfigure}
\caption{(a) A visualization of undersampled data when the proposed undersampling technique is applied to a simulated $3D$ dataset. The square empty points are original data and the triangular filled points are undersampled data. (b) A pictorial illustration of the proposed undersampling approach when data lies in the same plane.}
\end{figure}

\emph{Rejection criteria 2:} 
Once the first feature vector is found, look for the second feature vector $\mathbf{x}_j$, check if the angular displacement between the first selected feature vector $\mathbf{x}_i$ and the feature vector $\mathbf{x}_j$ w.r.t center $\mathbf{c}^k$ meets the threshold limit, i.e., if the condition $|\theta_{ci} - \theta_{cj}| > \delta \sigma_k $ is true, keep the vector or else check if the feature vectors lies in the same hyperoctant or orthant of the center $\mathbf{c}^k$ and $\mathbf{x}_i$.
If all three vectors $\mathbf{c}^k$, $\mathbf{x}_i$ and $\mathbf{x}_j$ are in the same plane then reject $\mathbf{x}_j$. Figure \ref{fig:under_sam1} illustrates an example of the case of comparing two feature vectors $\mathbf{x}_i$ and $\mathbf{x}_j$. It also possible that the angular distance between two feature vectors with respect to $\mathbf{c}^k$ is small, but if they are not lying in the same orthant then removing the feature vector may lead to loss of necessary information.      

The method of checking if two feature vectors lie on the same orthant is given as follows:
\begin{enumerate}
\item Find the difference between the cluster center $\mathbf{c}^k$ and the two feature vectors $\mathbf{x}_i$ and $\mathbf{x}_j$.
\item Set each element of the difference vector to either $1$  (if it is positive) or $0$ (if it is  negative).
\item Convert the binary string to decimal equivalent number. 
\item The decimal equivalent number provides the plane information of the feature vector w.r.t the center $\mathbf{c}^k$.
\end{enumerate}

\begin{algorithm}
\caption{Algorithm for under-sampling data}\label{undersam}
\footnotesize{
\begin{algorithmic}[1]
\Procedure{}{}
\State Run K-medoids and divide the majority class in K clusters with associated cluster centers $c_k$.
\State  Get the angular spread of each cluster and compute the approx number of data to be kept in each cluster.

\State Remove all scaled / repeated data points from each cluster
\State Compute the minimum threshold angle $\alpha_k$  between data points by which they are separated from each other.
\State Compute the error vectors by taking the difference between $c_k$ and the other data points in that cluster.

\State Obtain the Orthant of the error vectors.

\State Sort the data points based on their angular distance from $c_k$

\For {For each data point in each cluster}
\If {NOT $[\mathbf{\theta} (t + 1) - \mathbf{\theta} (t)] > \alpha_k]$ AND  $\mathbf{\theta} (t + 1)$ and $\mathbf{\theta} (t)$ are in same Orthant }
\State Discard the data point.
\EndIf
 \EndFor
 \EndProcedure
\end{algorithmic}}
\end{algorithm}

The hyper parameter $N_D$ can be kept as high as the total number of data in the majority class; i.e. the majority class is not under-sampled. In that case, the minority classes need to be over-sampled up to the level of the majority class. This may (or may not) result to a little increase in prediction accuracy but with huge computational cost if the classes are highly imbalanced. Moreover, the minority classes will be over crowded with less-informative data. On the other hand, if we select $N_D$ as small as the number of data in the next largest minority class, we may loose crucial information during under-sampling of the majority classes, which will degrade the prediction accuracy of the of algorithm. Hence, the choice of hyper parameter $N_D$ is a trade off between class recognition accuracy and the training time of the algorithm. It also depends on the imbalanceness of the classes. The best practice is to choose an intermediate value for $N_D$, which is lower than the number of data in the majority class but higher than the number of data in the next largest minority class and minority classes are over-sampled to that level. Here, we use cross validation to select the value for $N_D$.

\begin{figure}
\centering
\begin{subfigure}[b]{0.25\textwidth}
\includegraphics[scale=.08]{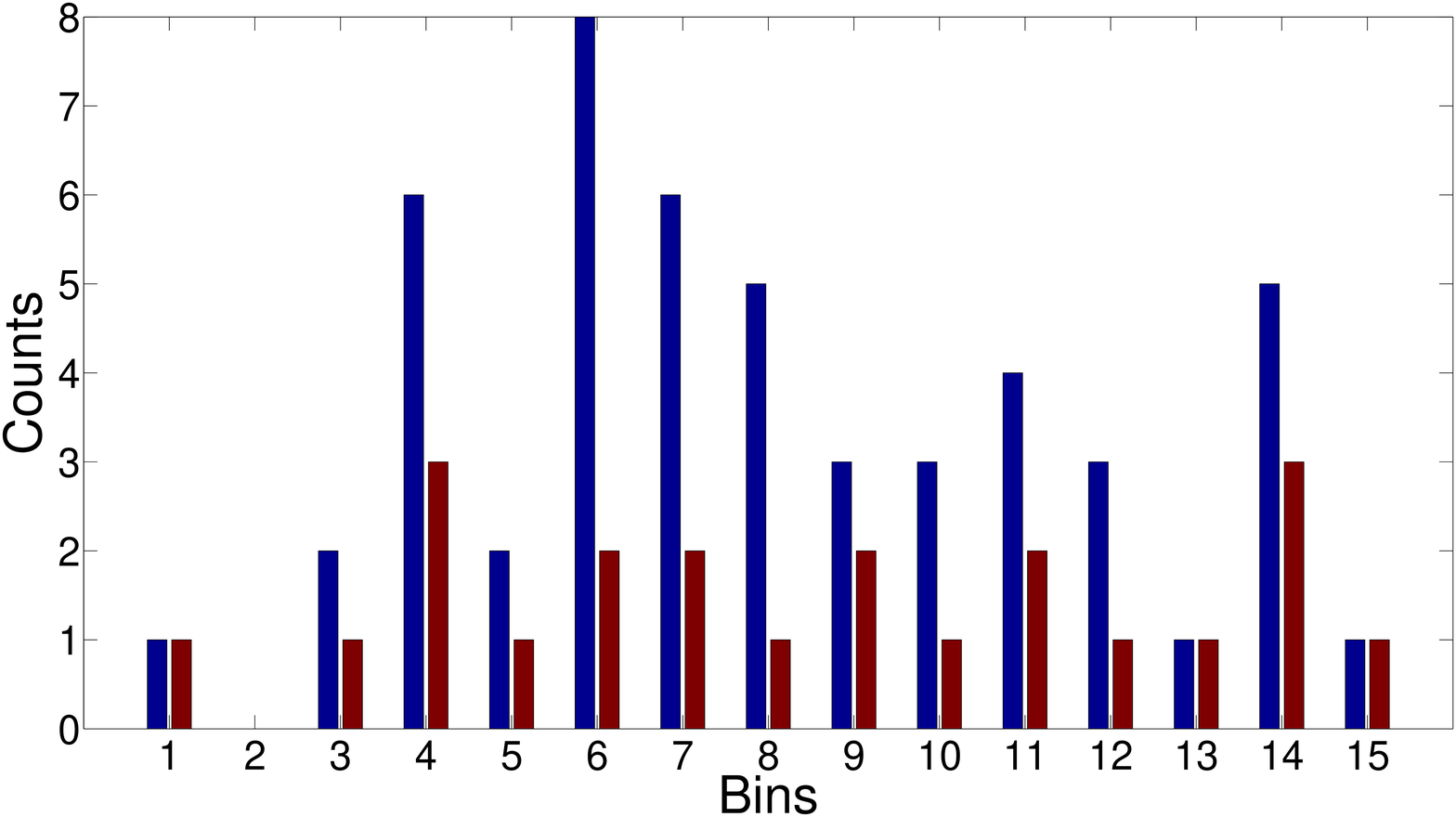}
\caption{}
\label{fig:hist_us_15}
\end{subfigure}~
\begin{subfigure}[b]{0.25\textwidth}
\centering
\includegraphics[scale=.08]{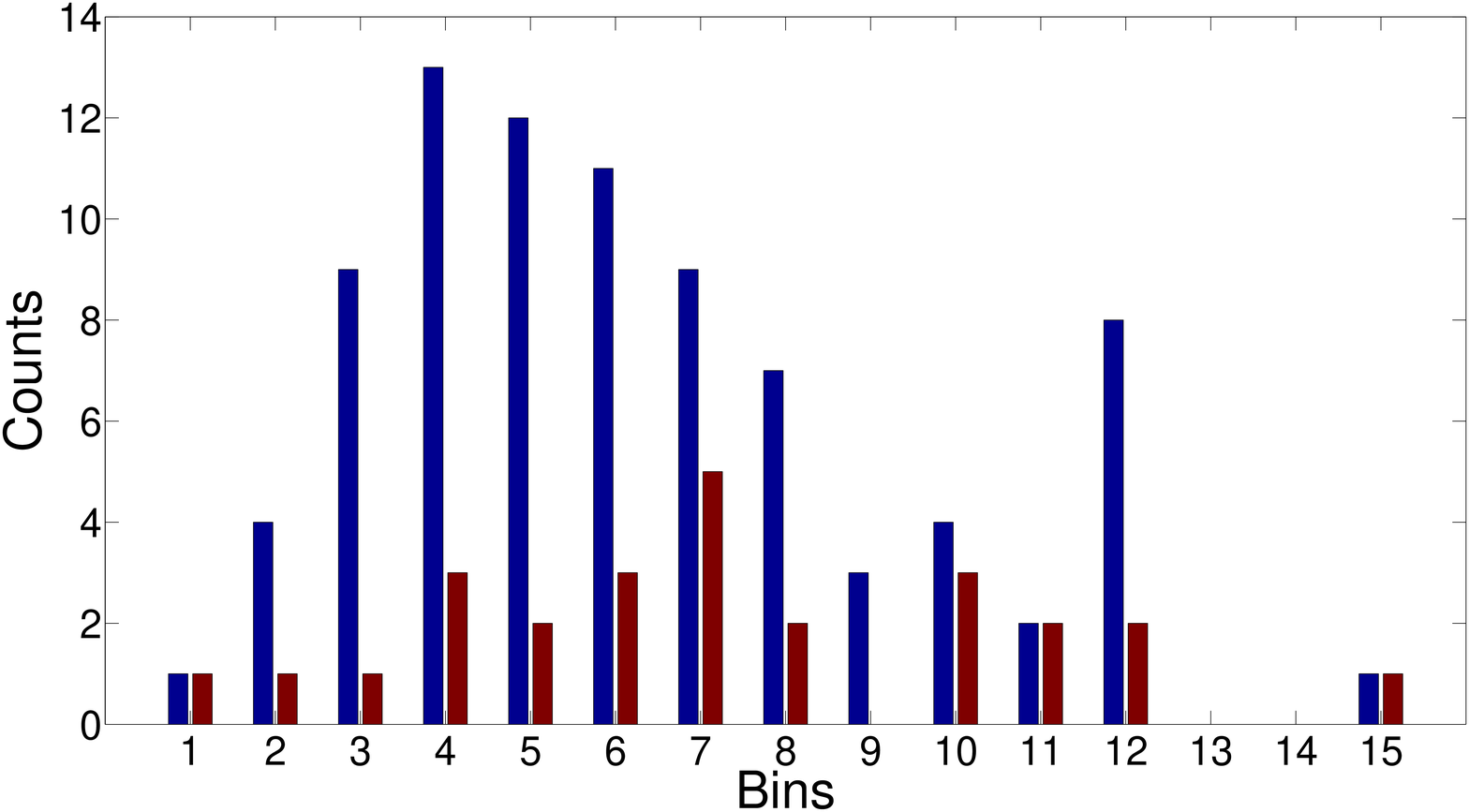}
\caption{}
\label{fig:hist_us_29}
\end{subfigure}
\caption{(a)Histogram plot of cluster 15 to show data distribution of different clusters after applying proposed undersampling. (b) Histogram plot of cluster 29 to show data distribution of different clusters after applying proposed undersampling.}
\label{fig:hist_us}
\end{figure}

The proposed undersampling algorithm is tested on a $3$ dimensional simulated data belong to a single class. The purpose of testing on the $3D$  dataset is to have a visual realization of the performance. Figure \ref{fig:3d_data_us} shows the performance of the proposed undersampling approach when applied to the simulated dataset. The squares are original data and the triangular points are undersampled data. For the dataset of size $2000$, we choose to select only $600$ data using the proposed undersampling approach. It can be observed that, the undersampled data still covers the entire class and maintains uniformity. It removes the unnecessary data from the dataset, yet keeps all the data which are sparsely distributed. Histogram plots of angular distances for two randomly chosen clusters are shown in Figures \ref{fig:hist_us_15} and \ref{fig:hist_us_29}. It is observed that the undersampled data in each of the clustered regions are almost uniformly scattered. The redundant or the less informative features are represented either by a single feature vector or by very few of them. It is observed from the histogram plots that the over crowded regions of the original class are made compact by selecting few feature samples. These selected features are the candidates from the crowded regions, which carry important characteristics of the entire class. It is also observed from the plots that no undersampled data is present in certain bins. Such cases can only happen when the data is either lying in the same orthant of the center and another feature vector which has already been selected previously that represents the data under consideration.    
\subsection{GICaPS oversampling approach}
\label{sec:oversam_des}

\begin{figure*}[htbp]
\centering
\begin{tabular}{ccc}
  \hspace{-0.7cm}\includegraphics[scale=0.25]{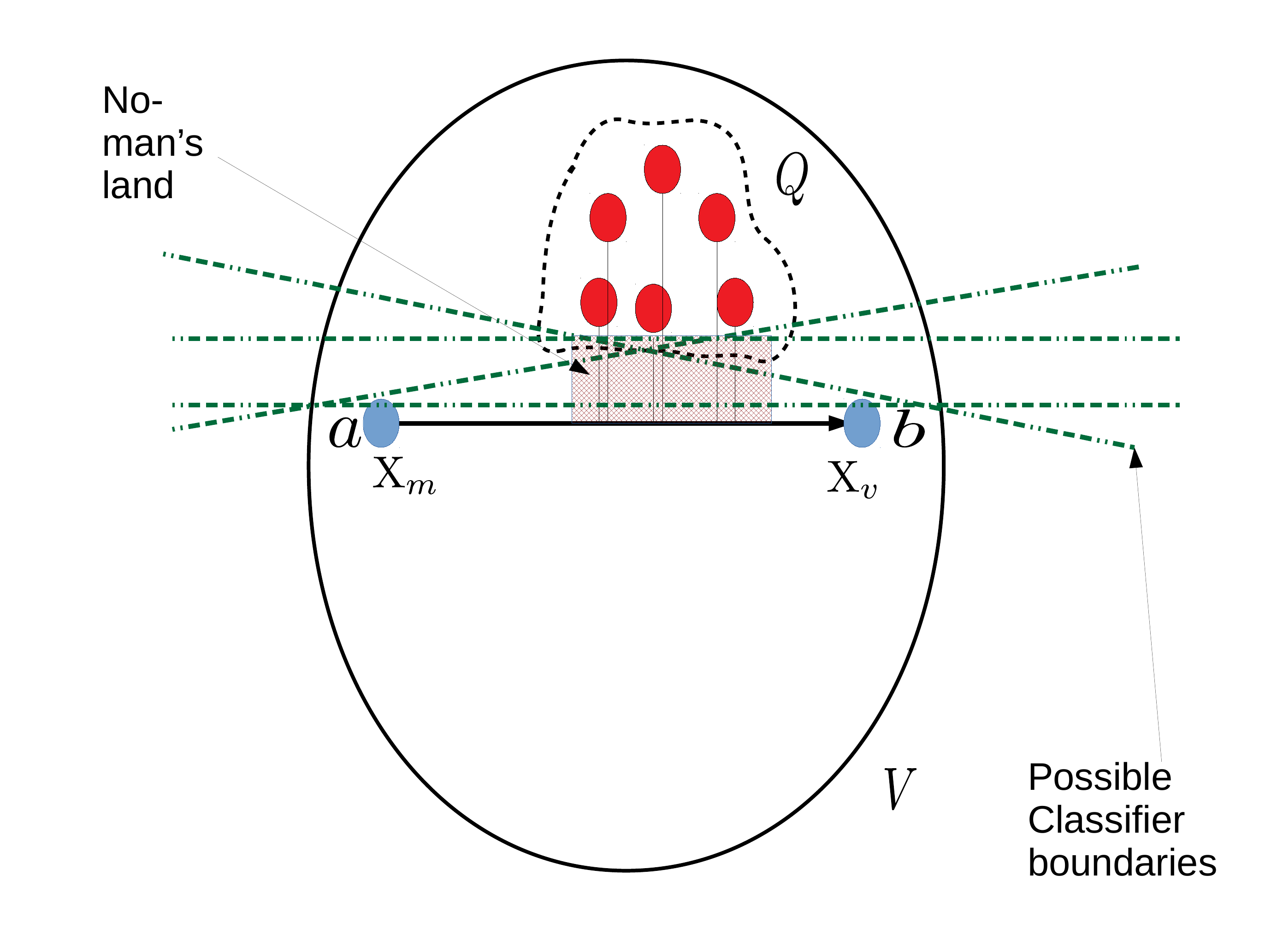} & \hspace{-2.3cm} 
  \includegraphics[scale=0.25]{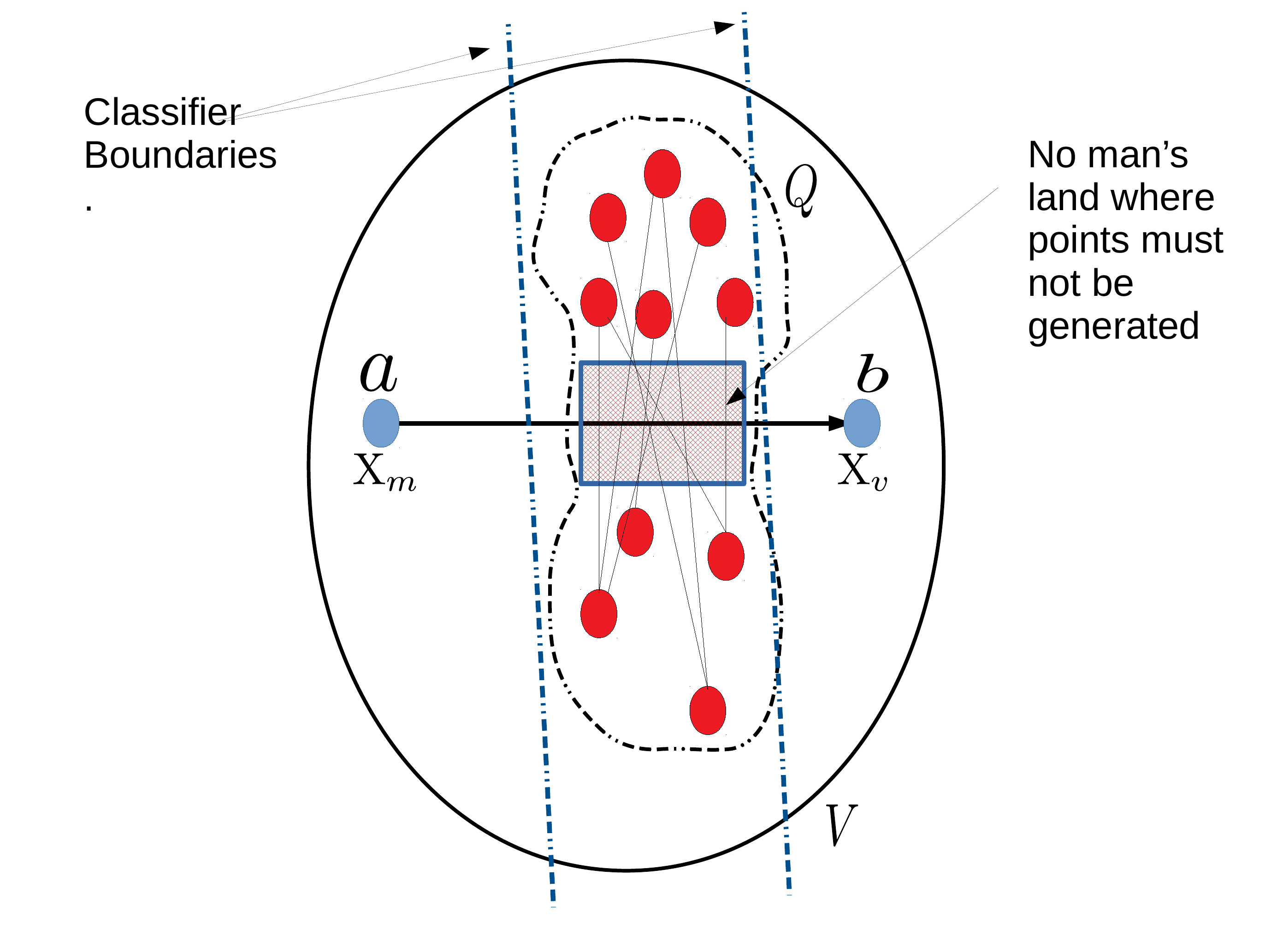} & \hspace{-2.3cm}
  \includegraphics[scale=0.25]{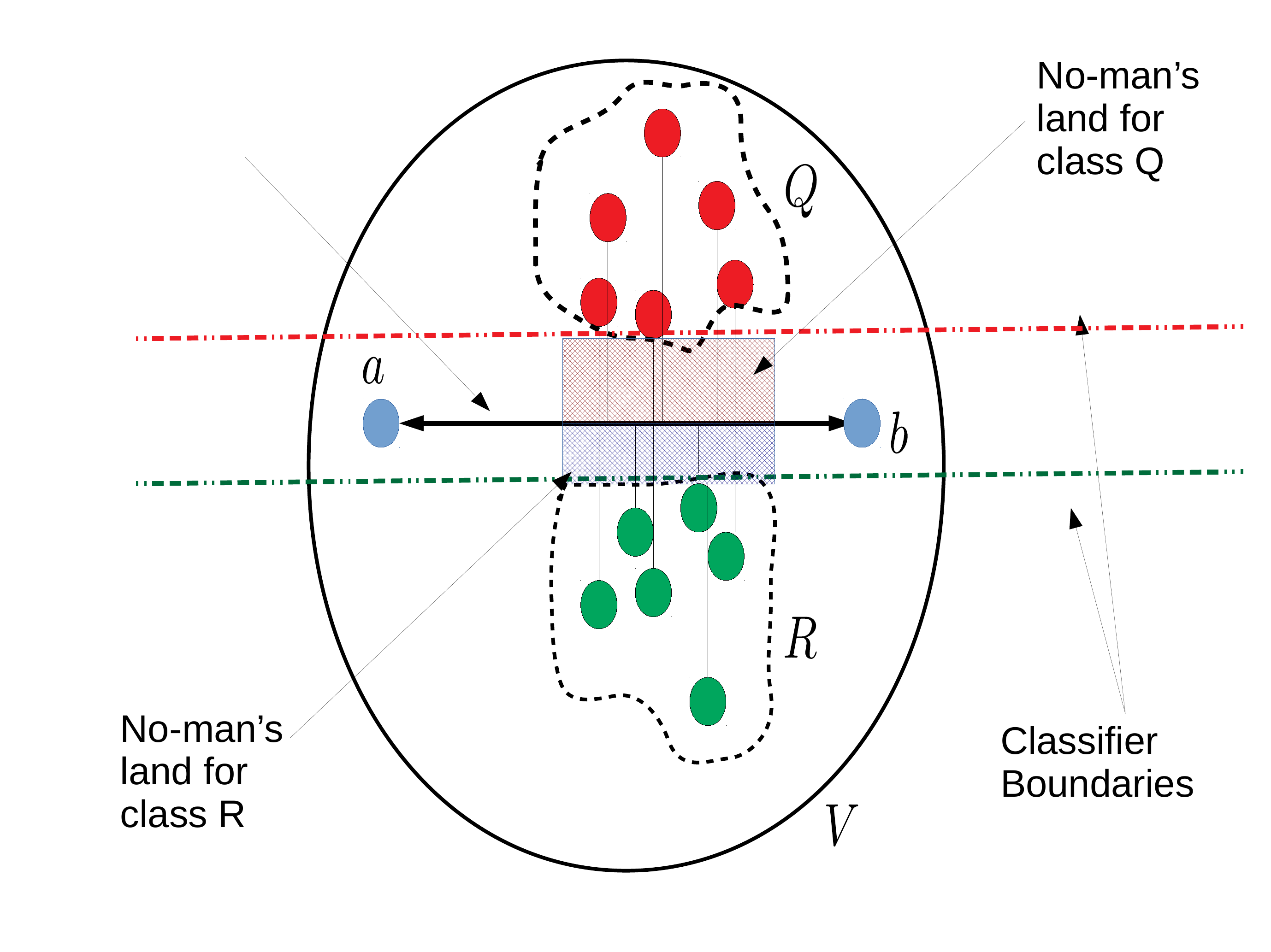} \\
  \hspace{-0.7cm}(a) & \hspace{-2.3cm}  (b)  & \hspace{-2.3cm}(c)
\end{tabular}
\caption{Defining \emph{no man's land} while generating new samples
between two existing data $a$ and $b$. (a) Case I: there are data points
belonging to a class Q lie on one side of the line joining data points
$a$ and $b$. (b) Case II: There are data points belonging to same
class Q on either side of the line joining data points $a$ and $b$.
(c) Case III: Data points belonging to two different classes Q and R
lie on either side of the line joining $a$ and $b$. Here it is assumed
that the data points $a$ and $b$ do not belong the class Q or R. }
\label{fig:nml_cases}
\end{figure*}

The oversampling algorithm presented here creates synthetic data samples for minority classes. The algorithm first checks the boundaries of the existing classes and then generates new data samples in the feasible regions. The proposed oversampling approach has the following features:
\begin{enumerate}
\item The proposed algorithm does not violate class boundaries while generating new synthetic samples. 
 The new data is generated in such a manner that it would not create confusion for the training module while learning the class distribution. In contrast, the existing algorithms such as SMOTE do not care about the inter class interference while interpolating new data. Other well established approaches, such as Borderline-SMOTE and ADASYN generate synthetic data only in the borderline of the minority classes.
 \item The proposed methodology generates new data samples between two data samples considering
the possible interference due to the data from neighborhood classes. (Please see Fig. \ref{fig:nml_cases}). 
\item The number of data to be generated in the minority class can be
defined by the user.
\end{enumerate}
The key idea of the proposed oversampling technique is that, while interpolating synthetic data between two points $\mathbf{x}_m$ and  $\mathbf{x}_v$, both belong to class $i$, we must consider the interference of other data points belonging to class $j$ with $j\neq i$.  This is because, the synthetic data may fall into other class's ambit if the criteria are not set properly. Such interference would essentially create confusion for the classifier to recognize class identity of a data point in that region. Considering the fact, we calculate the regions where such interference may occur. The regions are denoted as \textit{no man's land}. Any data interpolated in that region would be an \emph{illegal interpolation}. New synthetic data is generated avoiding the \emph{no man's land}. 

The concept of \emph{no man's land} can be better understood by
analyzing Figure \ref{fig:nml_cases} that shows three different
scenarios that one may come across while generating new data points
between two existing data points $a$ and $b$ within a given region,
say, $V$. The sub-figure (a) shows the first case where existing data
points belonging to a class Q may lie only on one side of the line
joining data points $a$ and $b$. The sub-figure (b) shows the case
where existing data points belonging to a class Q may lie on either
side of the line joining data points $a$ and $b$. The sub-figure (c)
shows the case where data points belonging to different classes Q and
R may lie on either side of the line $\overline{ab}$. Any other case can be
analyzed by combining these three cases. In each of these three cases,
the no-man's land represents the region that should be avoided while
generating data point between $a$ and $b$. Generating data points in
these (no-man's land) regions will disturb the classifier boundaries
between the classes. The exact shape of no-man's land will depend on
the distribution of data points within and between the classes.
However, for a two-dimensional dataset, it can be safely represented
by rectangles as shown in Figure \ref{fig:nml_cases}. The height $h$
of this rectangle could be minimum perpendicular distance of the
points of class Q or R from the line $\overline{ab}$. The length $l$ of
this rectangle includes all the points where the lines between the
points (of class Q) intersect with the line $\overline{ab}$. These points
of intersection may also be obtained by drawing projections from the
points of class Q as well, the former providing more conservative
estimate of the parameter $l$.
Considering all these above criteria we propose a mathematical approach for calculating \emph{no man's land} presented below in this section.  It is assumed that the dataset has total
$\mathbb{C}$  number of classes and the class $i$ is under
consideration in which new data needs to be interpolated. 
\begin{figure}
\centering
\captionsetup{width=1.0\linewidth}
\epsfig{file=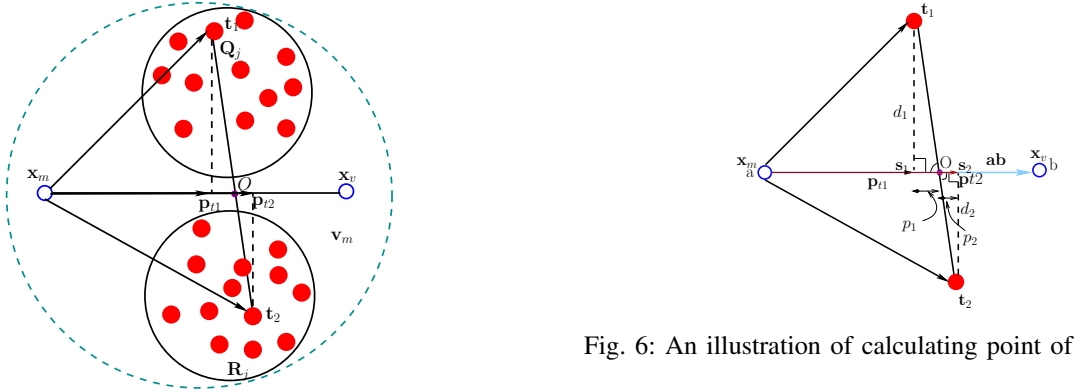,width=0.6\linewidth}
\caption{An overview of the proposed oversampling approach. Data points $\mathbf{x}_m$ and $\mathbf{x}_v$  belong to the class $i$ and the filled data points belong to other classes $j$. A region $\mathbf{V}_m$ contains the points $\mathbf{x}_m$ and $\mathbf{x}_v$ (between which synthetic data need to be generated) and the neighboring points of other classes $j$. }
\label{fig:oversam_ills}
\end{figure}
 
 Figure \ref{fig:oversam_ills} shows two neighboring data points $\mathbf{x}_m$ and $\mathbf{x}_v$  belong to the class $i$ and the rest of the filled data points belong to other classes. First, a region $\mathbf{V}_m$ is identified in the neighborhood of the point $\mathbf{x}_m$, where synthetic data samples need to be generated. Then the data points of class $j \in 1,2,...,\mathbb{C}, \ \ j\neq i$ are identified which fall in the region $\mathbf{V}_m$. Two regions $\mathbf{R}_j$ and $\mathbf{Q}_j$ are found where each connecting line from points in $\mathbf{R}_j$ to $\mathbf{Q}_j$ represents possible boundary of class $j$. The points in $\mathbf{R}_j$ are determined by a threshold criterion which defines the neighborhood of the pair $(\mathbf{x}_m, \mathbf{x}_v)$.  In the Figure \ref{fig:oversam_ills}, interpolation is to be done between data points $\mathbf{x}_m$ and $\mathbf{x}_v$. The points in $\mathbf{Q}_j$ are in fact the neighboring points of each data in $\mathbf{R}_j$. The filled circles within two regions $\mathbf{R}_j$ and $\mathbf{Q}_j$ are the data of other class $j$ in the region $\mathbf{V}_m$ which may interfere with the interpolated data between the data points $\mathbf{x}_m$ and $\mathbf{x}_v$. To avoid the interference during synthetic data generation, a method is proposed that selects legitimate places for data interpolation. Figure \ref{fig:oversam_proj} explains the method of finding the \textit{no man's land}.
Let's say, there is a data point $\mathbf{t}_1 \in \mathbf{Q}_j$ and a data point $\mathbf{t}_2 \in \mathbf{R}_j$ which are to be checked for any interference while interpolating the data along the vector $\mathbf{ab}$ connecting the  data points  $\mathbf{x}_m$ and $\mathbf{x}_v$. 
The steps are as follows:

\begin{figure}
\centering
\captionsetup{width=1.0\linewidth}
\epsfig{file=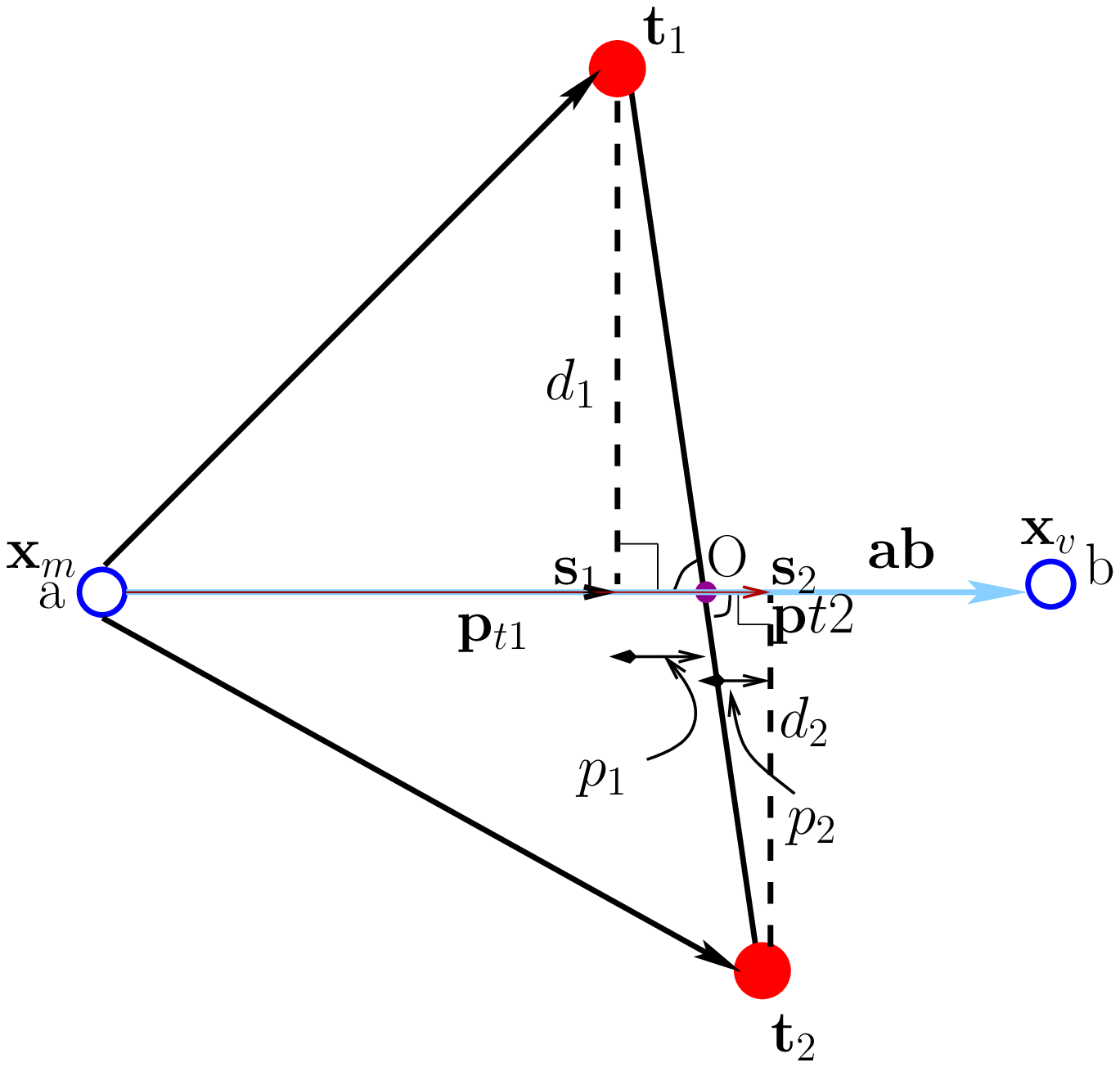,width=0.5\linewidth}
\caption{An illustration of calculating point of intersection.}
\label{fig:oversam_proj}
\end{figure}
The vector  $\mathbf{aO}$ is given as 
\begin{align}
\label{eq:proj}
 \mathbf{aO} = \mathbf{p}_{t1}  +  \mathbf{s_1O}
\end{align}
The triangles with points $\mathbf{t}_1$, $\mathbf{O}$, $\mathbf{s}_1$  and $\mathbf{t}_2$, $\mathbf{O}$, $\mathbf{s}_2$ forms two similar triangles. Thus, it can be written that the vectors $\mathbf{p}_{t1}$ and $\mathbf{p}_{t2}$ are the projections of the vectors $\mathbf{at}_1$ and $\mathbf{at}_2$ respectively on the line $\mathbf{ab}$. 
Therefore, following relations can be drawn. 
\begin{align}
 \frac{d_2}{d_1} = \frac{p_2}{p_1}\\
\frac{d_2+d_1}{d_1} =\frac{p_2+p_1}{p_1} \\
p_1 = \frac{p_2 + p_1}{d_2+d_1} \times d_1
\end{align}
where,
\begin{align}
 p_1 + p_2 = \Vert \mathbf{p}_{t2} -  \mathbf{p}_{t1} \Vert \\
d_1 = \Vert \mathbf{at}_1 - \mathbf{p}_{t1} \Vert  \\
d_2 = \Vert \mathbf{at}_2 - \mathbf{p}_{t2} \Vert \\
\mathbf{s_1O} =   (\mathbf{p}_{t2} -  \mathbf{p}_{t1}) \times \frac{d_1}{d_1 + d_2}
\end{align}
The equation \ref{eq:proj} can now be written as:
\begin{align}
\label{eq:intersecting_point}
\mathbf{aO}= \mathbf{p}_{t1} +   (\mathbf{p}_{t2} -  \mathbf{p}_{t1} )\times \frac{d_1}{d_1 + d_2}
\end{align}
The projections $\mathbf{p}_{t1}$ and $\mathbf{p}_{t2}$ can be calculated as
\begin{align}
 \mathbf{p}_{t1} = \frac{(\mathbf{ab})(\mathbf{ab})^T (\mathbf{at}_1) }{(\mathbf{ab})^T(\mathbf{ab}) }
\end{align}
\begin{align}
 \mathbf{p}_{t2} = \frac{(\mathbf{ab})(\mathbf{ab})^T (\mathbf{at}_2) }{(\mathbf{ab})^T(\mathbf{ab}) }
\end{align}
Let us also assume that the line connecting $\mathbf{t}_1$ and $\mathbf{t}_2$ intersects $\mathbf{ab}$ at point $\mathbf{O}$.
Till now, it was assumed that the crossing point $\mathbf{O}$ is in between $\mathbf{x}_m$ and $\mathbf{x}_v$. However, there might be cases where $\mathbf{O}$ might fall outside $\mathbf{ab}$. Hence, the next step is to check if the point  $\mathbf{O}$ lies on or near to the vector $\mathbf{ab}$. 
To check if the line joining the vectors $\mathbf{t}_1$ and $\mathbf{t}_2$ crosses the intersecting point $\mathbf{O}$ following verification is done.
\begin{align}
\label{eq:chk_lyes}
\begin{split}
\Vert \mathbf{aO} \Vert < \Vert \mathbf{ab} \Vert \\
 \mbox{and} \\
 \Vert \mathbf{ab} - \mathbf{aO} \Vert < \mathbf{ab}
 \end{split}
\end{align}
If the above conditions are satisfied, then the \emph{crossing distance} needs to be calculated related to point $\mathbf{t}_1$ and $\mathbf{t}_2$. The \textit{crossing distance} is defined as the shortest distance of the line joining $\mathbf{t}_1$ and $\mathbf{t}_2$ from point $\mathbf{O}$. 
Let us define
\begin{align}
\mathbf{Ot}_1 = \mathbf{aO} - \mathbf{at}_1\\
\mathbf{at}_{12} = \mathbf{at}_2 -  \mathbf{at}_1.
\end{align}
The \emph{crossing distance} is thus calculated as:
\begin{align}
\label{eq:cross_dist}
C_{dist} = \Vert \mathbf{Ot}_1 \Vert \times \sin(\theta_0)
\end{align}
where, $\theta_0$ is the angle between the vectors $\mathbf{Ot}_1$ and $\mathbf{at}_{12}$. 
The \emph{no man's land} is estimated based on the above calculation. A threshold is set to find the region of \emph{no man's land}. If $C_{dist}$ is less than a threshold value, then no data is interpolated within the region. The region thus comes under \emph{no man's land}. 
The intersecting points and the \emph{crossing distances} are calculated for all the points within the regions $\mathbf{R}_j$ and $\mathbf{Q}_j$. And the same process is repeated for all the $j \in (1, \ldots, \mathbb{C})$ classes. Thus, \emph{no man's land} between $\mathbf{x}_m$ and $\mathbf{x}_v$ is identified by calculating all the intersections on $\mathbf{ab}$ for $\mathbb{C}$ classes in the neighborhood. First, the intersections $\mathbf{O}_1, \ \mathbf{O}_2,  \ ..., \ \mathbf{O}_K$ ($K$ is the total number of data samples from the other classes that cause interference) are identified using the procedure explained above. The closest and farthest $\mathbf{O}_k$ to $\mathbf{x}_m$ define the range of \emph{no man's land}.
Once the \emph{no man's land} is identified, the number of new points need to be generated for each data $\mathbf{x}_m$ in the class $i$ is calculated next. 
The free space between the vector $\mathbf{x}_m$ and $\mathbf{x}_v$ is given as:  
\begin{align}
S^{\mathbf{x}_m \mathbf{x}_v} = \rho \{ \Vert \mathbf{ab}  \Vert - \Vert \mathbf{O}_{max} - \mathbf{O}_{min}\Vert \}
\end{align}
where, $\mathbf{O}_{max}$ and $\mathbf{O}_{min}$ are the longest and shortest vector from $\mathbf{x}_m$ to $\mathbf{O}_k$ (farthest and closest intersection). The effect of $\rho$ is to increase the range of the \emph{no man's land}, such that the region does not start and end strictly at closest and farthest $\mathbf{O}_k$ respectively.
Total number of data to be interpolated within the $\mathbf{V}_m$ region is given as:
\begin{align}
N_{\mathbf{x}_m} ^{\mathbf{V}_m} = \frac {   H_i  \sum_{\mathbf{V}_m} S^{\mathbf{x}_m \mathbf{x}_v} }{ \sum_{\mathbf{x}_m} \sum_{\mathbf{V}_m} S^{\mathbf{x}_m \mathbf{x}_v}}
\end{align}
and the number of data to be interpolated in between two data points ($\mathbf{x}_m$ and $\mathbf{x}_v$) is given as: 
\begin{align}
N_{\mathbf{x}_m} ^{\mathbf{x}_v} = \frac {  N_{\mathbf{x}_m} ^{\mathbf{V}_m}  S^{\mathbf{x}_m \mathbf{x}_v} }   { \sum_{\mathbf{V}_m} S^{\mathbf{x}_m \mathbf{x}_v}}
\end{align}
The method of interpolating $N_{\mathbf{x}_m}^{\mathbf{x}_v}$ number of data within the free regions of the line $\mathbf{ab}$ for the data vector $\mathbf{x}_m$ is given by: 
\begin{align}
\mathbf{x}_m^{new} = \mathbf{x}_m + \gamma \frac{\mathbf{x}_v - \mathbf{x}_m}{N_{\mathbf{x}_m}^{\mathbf{x}_v}} + \mathbf{r}_m 
\end{align}
where $\gamma = 1, 2, 3, \ldots, N_{\mathbf{x}_m}^{\mathbf{x}_v}$ and $\mathbf{r}_m$ is a small random noise.
\begin{algorithm}
\caption{Algorithm for Oversampling data}\label{oversam}
\footnotesize{\begin{algorithmic}[1]
\Procedure{}{Given $\mathbb{C}$ classes in a dataset and class $\mathbf{c}_i$ is under consideration for interpolation. Let us assume that data has to be interpolated between two points $\mathbf{x}_m$ and $\mathbf{x}_v$. }
\For {class $\mathbf{c}_i$ to $\mathbb{C}$ }
\For {each data $\mathbf{x}_m$ in the class $\mathbf{c}_i$ }

\State Identify region $\mathbf{V}_m$ that includes the data points $\mathbf{x}_m$ and $\mathbf{x}_v$ from class $\mathbf{c}_i$ and  neighboring points from all other classes $\mathbf{c}_j$ where $j = 1, \ldots, \mathbb{C}$ and $j\neq i$. 
\For {each vector $\mathbf{x}_v$ in the region $\mathbf{V}_m$}

\State Find two regions $\mathbf{R}_j$ and $\mathbf{Q}_j$ near to the line  joining the points  $\mathbf{x}_m$ and $\mathbf{x}_v$ representing possible boundary of class $j$.


\State To avoid interference while interpolation select legitimate places for data interpolation between two points.

\State Find \emph{No man's land} using the method explained in Section \ref{sec:oversam_des}. 

\For {For each data point in the region $\mathbf{R}_j$ and $\mathbf{Q}_j$}
\State Calculate the intersecting points and crossing distance using the technique described in Section \ref{sec:oversam_des}.
\State repeat the above step for all class $j = 1, \ldots, \mathbb{C}$ and $j\neq i$.
 \EndFor

\State After \emph{No man's lands} are identified, calculate the total number of points to be between for each data points $\mathbf{x}_m$ within the region $\mathbf{V}_m$.
\State Calculate the empty region $ S^{\mathbf{x}_m \mathbf{x}_v} $ between $\mathbf{x}_m$ and $\mathbf{x}_v$  
\State Find $S^{\mathbf{x}_m \mathbf{x}_v}$ for all the points in the region $\mathbf{V}_m$.
\EndFor   
  \State Calculate the space vector for each data point $\mathbf{x}_m$ in the class $\mathbf{c}_i$. 
    \State Calculate the total number of data to be interpolated in the neighborhood of the vector $\mathbf{x}_m$ 
 \EndFor  
\EndFor 

 \EndProcedure
\end{algorithmic}}
\end{algorithm}
\begin{figure}
 \epsfig{file=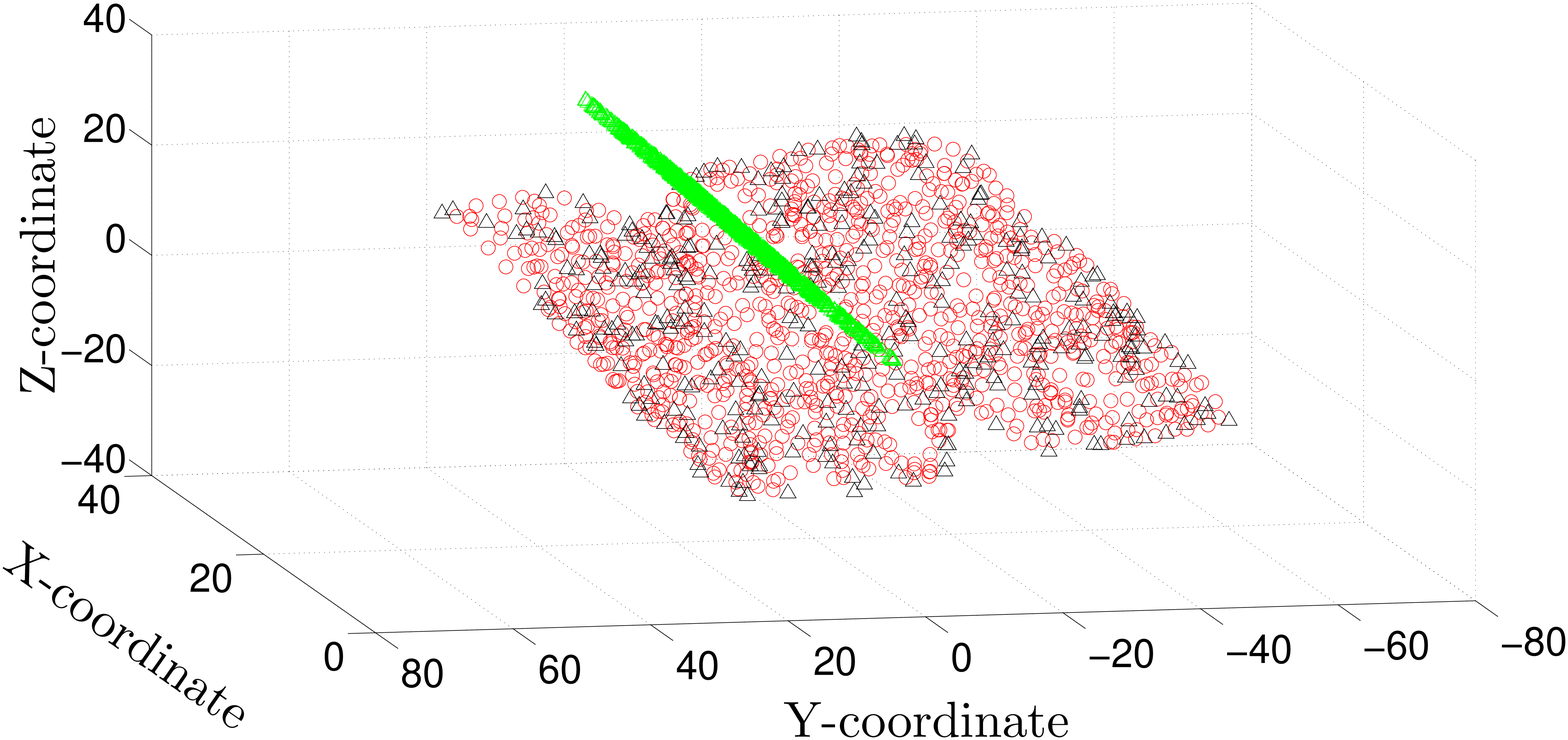,width=1.0\linewidth}
 \captionsetup{width=0.8\linewidth}
\caption{Oversampling results of proposed approach for a synthetically generated $3$ dimensional dataset having two different class. Class $1$ view}
\label{fig:oversam_class1_view}
\end{figure}
\begin{figure}
  \epsfig{file=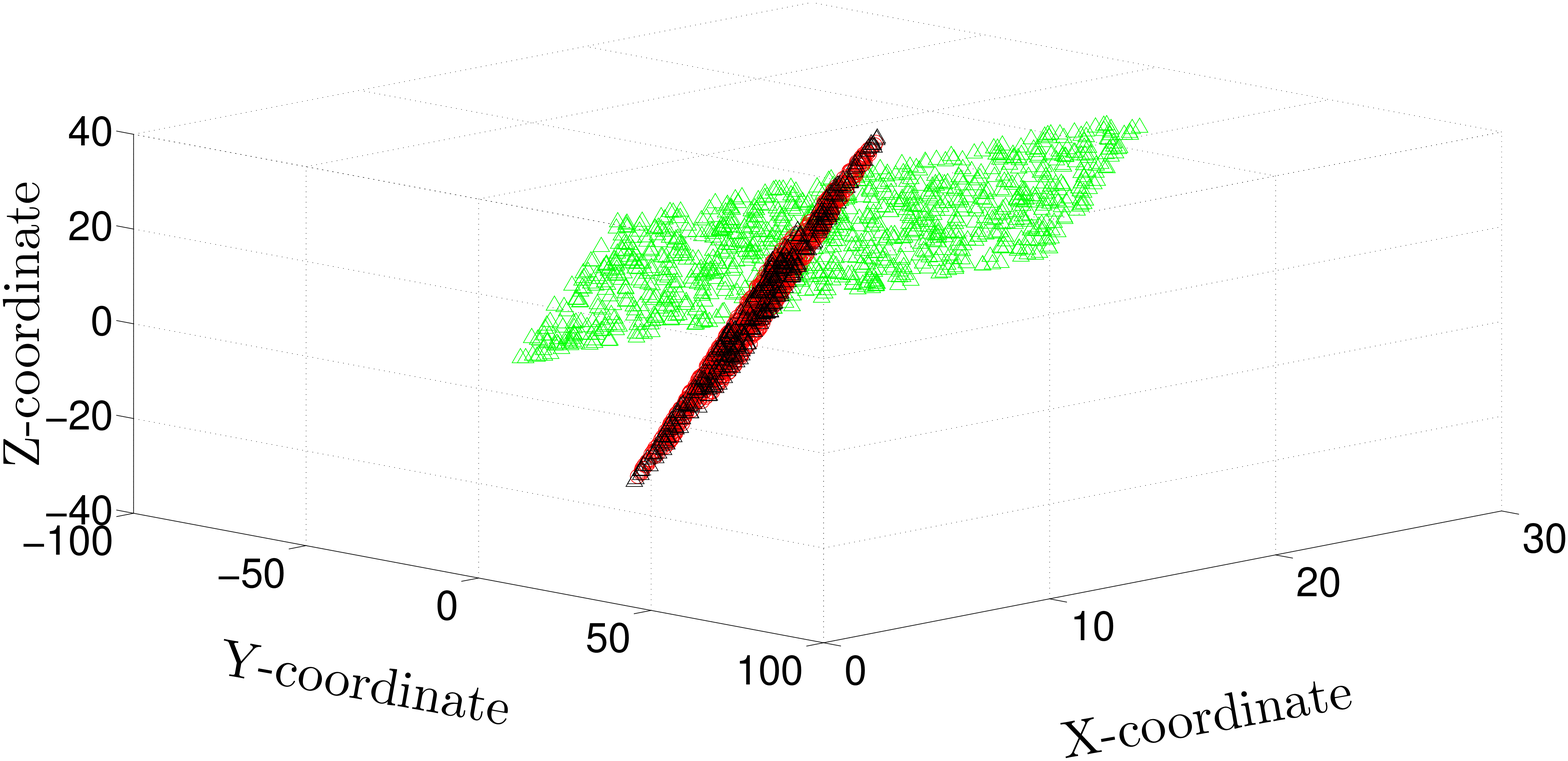,width=1.0\linewidth}
 \captionsetup{width=0.8\linewidth}
\caption{Oversampling results of proposed approach for a synthetically generated $3$ dimensional dataset having two different classes. Class $2$ view}
\label{fig:oversam_class2_view}
\end{figure}
\begin{figure}
  \epsfig{file=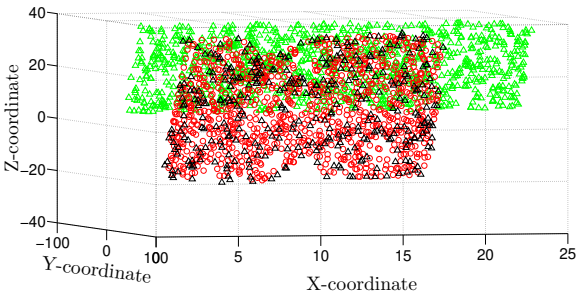,width=1.0\linewidth}
   \captionsetup{width=0.8\linewidth}
\caption{Oversampling results of proposed approach for a synthetically generated $3$ dimensional dataset having two different classes. Overall view}
\label{fig:oversam_both_class_view}
\end{figure}
\subsection{Illustration with a simulated dataset}
\begin{figure}
 \epsfig{file=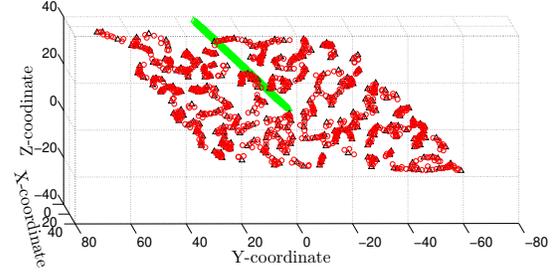,width=0.9\linewidth}
\captionsetup{width=0.95\linewidth}
\caption{Oversampling results of a synthetically generated two class dataset using SMOTE.}
\label{fig:oversam_smote_syn}
\end{figure}

\begin{figure}
 \epsfig{file=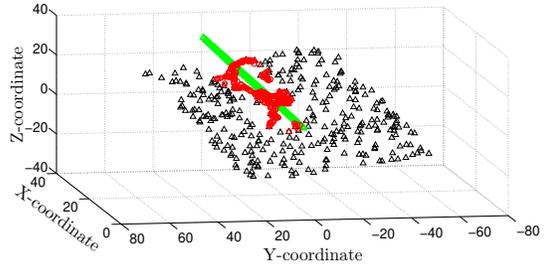,width=0.9\linewidth}
\captionsetup{width=0.95\linewidth}
\caption{Oversampling results of a synthetically generated two class dataset using ADASYN.}
\label{fig:oversam_adasyn_syn}
\end{figure}

A two class simulated dataset is created to visualize and compare the performance of the GICaPS oversampling approach with existing well established techniques, such as SMOTE and ADASYN. The results of SMOTE and ADASYN are shown in Figure \ref{fig:oversam_smote_syn} and Figure \ref{fig:oversam_adasyn_syn} respectively and the performance of the GICaPS oversampling approach is already shown in Figure \ref{fig:oversam_class1_view}. The black triangular points are the original data of minority class and red circles are the synthetically generated data. The green triangular points are the original data of majority class. 
It can be observed that, the distribution of the synthetically generated data is not uniform in both SMOTE and ADASYN. SMOTE interpolates data between the two existing points without even considering the interference of the majority class. Also, as SMOTE randomly decides which of the K nearest neighbor is to be selected for interpolation, it may happen that more relevant point gets missed and too many data gets interpolated between two closely placed data points. Figure \ref{fig:oversam_adasyn_syn} shows the results of synthetic data generation for the minority class data. The observation clearly shows that the new data is generated only near to the border. Even most of the synthetically generated data falls on the territory of the majority class, which will definitely mislead the classifier. The observation gives a clear illustration that the performance of ADASYN is even worse than SMOTE. In contrast to both, the proposed approach takes every neighboring point into consideration, and equal spacing is maintained between all the points while interpolating data. Moreover, unlike SMOTE and ADASYN, the GICaPS decides how many points are to be interpolated between any neighboring points. It also decides whether to interpolate data between two points or not, based on possibility of conflict with other class boundaries.
A visualization of the GICaPS oversampling approach is shown in Figure \ref{fig:oversam_class1_view}.

Table \ref{tab:oversam_perf_evaluation} shows an ablation study performed to proof the efficacy of the GICaPS-Oversampling approach. Two randomly chosen class data are used for this purpose. The chosen minority class is oversampled using GICAPS-oversampling approach, SMOTE and ADASYN. Support vector machine is used to perform this experiment. Minimum distance between support vectors of two classes is calculated for each of the approaches and the same is presented in the Table \ref{tab:oversam_perf_evaluation}. We have shown five different datasets for this ablation study. The statistical analysis clearly shows that the proposed approach maintains maximum margin between two classes after oversampling is performed.  
\begin{table}
\centering
\captionsetup{width=1.0\linewidth}
\caption{A statistical evaluation of class boundary using minimum euclidean distance between support vectors of two different classes chosen randomly. The performance are also compared with state-of-the-art techniques: SMOTE and ADASYN. Experiments are performed on five challenging datasets, namely Ionosphere, Shuttle, Glass, Abalone, Indian Diabetic and Shuttle. It is to be noted that oversampling techniques are applied to at least one of the chosen classes.}
\label{tab:oversam_perf_evaluation}
\begin{tabular}{ |p{1.4cm}|p{1.6cm}|p{1.5cm}|p{1.5cm}| }
 \hline
 \multicolumn{4}{|c|}{A statistical evaluation of class boundary.} \\
 \hline
 Dataset & Methods & \# PCA   & Min distance \\
 & & component & \\
 \hline
    Glass & GICaPS-O  &4&\bf{2.0297}    \\
 \cline{2-3}
  
      & SMOTE    &4 &0.4916\\
   \cline{2-3}
      & ADASYN    & 4 &  0.4974\\
  \hline 
    Diabetic  & GICaPS-O  &4&\bf{0.2224}    \\
 \cline{2-3}
  
      & SMOTE    &4 &0.1358  \\
   \cline{2-3}
      & ADASYN    & 4 &  0.1962 \\
  \hline 
    Abalone  & GICaPS-O  &6&\bf{0.5248}    \\
 \cline{2-3}
  
      & SMOTE    &4 &0.0530  \\
   \cline{2-3}
      & ADASYN    & 4 &  0.0759 \\
  \hline 
 
    Shuttle  & GICaPS-O  &4&\bf{0.7968}    \\
 \cline{2-3}
  
      & SMOTE    &4 &0.0  \\
   \cline{2-3}
      & ADASYN    & 4 &  0.3053  \\
  \hline
 Ionosphere  & GICaPS-O  &4&\bf{0.5372}    \\
 \cline{2-3}
  
      & SMOTE    &4 &0.3526  \\
   \cline{2-3}
      & ADASYN    & 4 &  0.2087  \\
  \hline
\end{tabular}
\end{table}
It is to be noted that, angular information can also be used for the proposed oversampling approach. The main intension of using the angular information in the undersampling approach to remove redundant data and to avoid removal of more informative data which may not be taken care when the Euclidean distance considered as the rejection criteria. As, oversampling involves synthetic data generation, we opted to use Euclidean distance for the shake of lesser computational complexity. However, one can use angular information that will avoid generation of redundant data.
\section {Recognition using mixture of Gaussians}
\label{subsec:pain_gmm}
Gaussian distribution has a wide range of applicability in realistic distributions. The performance and applicability of the estimating model is further enhanced when multiple Gaussians are used in place of one to model the data distribution. In this work, the distribution of the training data is captured using Gaussian mixture model which is a linear combination of finite number of Gaussians. The recognition problem is solved as a regression problem. The class identity is predicted by the regressive model, created by the mixture of Gaussians. We select regression over the standard GMM classification because of two reasons: i) The execution time is faster when the class dimension is high and ii) we want to show that a regressive model can also perform well with the dataset created using the proposed data balancing technique.
\subsection {Recognition model using mixture of Gaussians}
\label{subsec:gmm}
The regressive model for the recognition problem is given in the following:
\begin{align}
 y = f(\mathbf{x})
\end{align}
where, $\mathbf{x}\in \Re^D$ is the feature vector and $y\in \Re$ is the class labels and can take values $y_i=\mathcal{I}_i, \ i=1,2...\mathbb{C}$, where $\mathbb{C}$ is the total number of class labels.
Let's assume that the random feature vector $\mathbf{x}$ can be matched with a class variable $y$ and the joint probability density can be modeled using the mixture of Gaussians \cite{bishop2006pattern}. The probability distribution of the random variable $\boldsymbol{\xi}=[\mathbf{x};y]$ fits into the GMM and is given by
\begin{align}
 p(\boldsymbol{\xi}) &= \displaystyle\sum_{k=1}^K \pi_k\mathcal{N}(\boldsymbol{\xi};\mu_k, \Sigma_k) \\
 & = \displaystyle\sum_{k=1}^K \pi_k \frac{1}{\sqrt{(2\pi)^{D+1}|\Sigma_k|}}e^{\frac{1}{2}(\boldsymbol{\xi}-\mu_k)^T\Sigma_k^{-1}(\boldsymbol{\xi}-\mu_k)}
\end{align}
 where, $\pi$ is the class prior or prior probability and $\mathcal{N}(\mu_k, \Sigma_k)$ is the $k^{th}$ Gaussian distribution with $\mu_k$ being the mean and $\Sigma_k$ is the co-variance of the distribution 
 and is given by:
  \begin{align}
\label{eq:mu_sigma_f}
  \mu^k=\begin{bmatrix}\mu^{\mathbf{x}}_k \\ \mu^{y}_k\end{bmatrix} \hspace{3pt}and \hspace{5 pt}  \Sigma^k=\begin{bmatrix}
\Sigma^{\mathbf{x}}_k  & \Sigma^{\mathbf{x}y}_k \\ 
\Sigma^{{\mathbf{x}y}}_k & \Sigma^{y}_k
\end{bmatrix}.
\end{align}
The posterior  $p(y|\mathbf{x})$ for a given feature vector $\mathbf{x}$ and component $k$ can be found using Gaussian mixture regression. The posterior mean estimate $\hat{y}$ can be found as 
\begin{align}
 \hat{y} = \displaystyle\sum_{k=1}^K h(k)\left [ \mu^y_k + \Sigma_k^{\mathbf{x}y} (\Sigma_k^{\mathbf{x}})^{-1}(\mathbf{x} - \mu^{\mathbf{x}}_k ) \right ]
\end{align}
where, $h(k) = \frac{p(k)p(\mathbf{x}|k)}{\Sigma_{k=1}^K p(k)p(\mathbf{x}|k}$. The class variable $y$ is given by
$y=g(\hat{y})$, where, $g:\Re  \to \Re \quad \forall \mathbf{x} \in \Re^D$ maps $\hat{y}$ to its nearest class value.
The parameters of the Gaussian distributions are estimated using Expectation maximization (EM), since the maximum likelihood does not work here as there is no closed form solution for GMM. The EM algorithm can be found in \cite{bishop2006pattern}.

\section{Experimental Results and Discussions}
\label{sec:exp_res}
The proposed algorithm has been tested on ten popular imbalanced datasets. The datasets are chosen in such a manner that it contains numeric attributes and no missing data. Unlike other existing imbalanced data handling techniques, this work includes multi-class datasets with number of classes as high as $23$ in case of abalone dataset and $15$ in case of UNBC-McMaster Shoulder Pain Expression Archive database \cite{lucey2011painful}.   
Classes containing only one instance have been removed from the datasets as the proposed data handing technique in its current state cannot generate new data with only one sample. In case of pain dataset \cite{lucey2011painful}, geometric features vector $\mathbf{x}\in\mathbb{R}^{22}$ is extracted from all the images present in the database. 
Initially the face is detected using Viola Jones' face detection algorithm \cite{viola2004robust} followed by Viola Jones' algorithm to detect two eyes. The centers of the eyes are detected using our propose approach presented in \cite{Majumder20141282}. We further calculate the rotation angle using center of two eyes. The face is then rotated to frontal face image using rotation transformation matrix. Geometric features are extracted from the normalized face images. The methods of geometric features extraction is given in \cite{Majumder20141282}. 

GICaPS undersampling is applied to only those majority classes in which the number of instances are much high. For example, the pain dataset contains 39835 samples in the majority class whereas, least number of data among minority classes is only 5. Spambase on the other-hand has 2788 instances in the majority class in contrast to 1813 samples in minority class. Another dataset, named shuttle has 7 different classes with number of instances in majority class as high as 34108 and the least number of data in minority class is only 6. In such cases, the rejected data from the majority class are included in test set. 
\begin{center}
\begin{table}
\centering\footnotesize{
\captionsetup{width=0.95\linewidth}
\caption{Data distribution of pain db after applying SMOTE. (The terms are: I-Intensity, DS-Data size)} 
\begin{tabular}{c c c c c c c c c} 
\hline\hline 
I & $I_0$ & $I_1$ & $I_2$ & $I_3$ & $I_4$ & $I_5$ & $I_6$ & $I_7$  \\  
DS & 39835 & 11632 & 9396 & 5636 & 3208 & 968 & 1080 & 212 \\
 \hline 
I & $I_8$ & $I_9$ & $I_{10}$ & $I_{11}$ & $I_{12}$ & $I_{13}$ & $I_{14}$ & $I_{15}$\\
DS & 316 & 128 & 268 & 304 & 192 & 88 & - & 20\\
\hline 
\label{tab:pain_data_dist_smote}
\end{tabular}}
\end{table}
\end{center}

The training and testing are done using $10$ fold cross validation technique. To validate the performance of the proposed data balancing technique, we compare the results with well established data handling techniques, such as SMOTE and ADASYN. It has been observed that, unlike ADASYN and GICaPS, the data distribution among classes after applying SMOTE remains skewed in most of the highly imbalanced datasets. Such an instance can be shown using the pain dataset. The distribution of data among classes after applying SMOTE, ADASYN and GICaPS are presented in Table \ref{tab:pain_data_dist_smote}, Table \ref{tab:pain_data_dist_adasyn} and Table \ref{tab:pain_data_dist_prop_app} respectively. In case of ADASYN, we have randomly picked $10000$ data from the majority class and for all the minority classes equivalent number of data are generated. This is done to reduce huge computational cost during training of the classification model. Source code of the proposed undersampling and oversampling approaches is available online \cite{sam_code}.
\begin{center}
\begin{table}
\centering\footnotesize{
\captionsetup{width=0.95\linewidth}
\caption{Data distribution of pain db after applying GICaPS (The terms are: I-Intensity, DS-Data size)} 
\begin{tabular}{c c c c c c c c } 
\hline\hline 
I & $I_0$ & $I_1$ & $I_2$ & $I_3$ & $I_4$ & $I_5$ & $I_6$   \\  
DS & 13502 & 13502 & 13502 & 13502 & 13502 & 13502 & 13502 \\
 \hline 
I& $I_7$ &$I_8$ & $I_9$ & $I_{10}$ & $I_{11}$ & $I_{12}$ & $I_{13}$  \\
 
DS & 13502 & 13502 & 13502 & 13502 & 13502 & 13502 & 13502 \\
\hline
I & $I_{14}$& $I_{15}$ & & & & &\\

 DS & - & 13502& & & & &\\
\hline 
\label{tab:pain_data_dist_prop_app}
\end{tabular}}
\end{table}
\end{center}

\begin{center}
\begin{table}
\centering
\footnotesize{
\captionsetup{width=0.95\linewidth}
\caption{Data distribution of pain db after applying ADASYN. (The terms are: I-Intensity, DS-Data size)} 
 \begin{tabular}{c c c c c c c c c} 
\hline\hline 
I & $I_0$ & $I_1$ & $I_2$ & $I_3$ & $I_4$ & $I_5$ & $I_6$ & $I_7$  \\  
DS & 10000 & 9958 & 9958 & 9958 & 9958 & 9958 & 9958 & 9958 \\
\hline 
I & $I_8$ & $I_9$ & $I_{10}$ & $I_{11}$ & $I_{12}$ & $I_{13}$ & $I_{14}$ & $I_{15}$\\
DS & 9958 & 9958 & 9958 & 9958 & 9958 & 9958 & \text{-} & 9958\\
\hline 
\label{tab:pain_data_dist_adasyn}
\end{tabular}}
\end{table}
\end{center}
The balanced datasets are trained using the GMR model. The recognition performances of all the datasets, in terms of overall accuracy, precision, recall, F-measure and G-Mean are presented in a tabular form as given in the Table \ref{tab:perf_evaluation}. The observation shows that, performance of GICaPS is significantly better than ADASYN and SMOTE in almost all the datasets. For instance, in case of pain database, average recognition accuracies of $13.197\%$ and $90.24\%$ are achieved for ADASYN and SMOTE respectively. Whereas, the recognition accuracy of the proposed data handling approach outperforms the recognition accuracy of both ADASYN and SMOTE. The proposed data balancing approach gives an average recognition accuracy of $98.81\%$ which is a significant improvement over SMOTE and ADASYN. The performance	of GICaPS is also compared with some of the recent state-of-the-art techniques, such as SIMO \cite{piri2018synthetic}, WSIMO \cite{piri2018synthetic}, SWIM\cite{sharma2018synthetic} and MOCAS (NN) \cite{lin2018minority}. The statistical comparisons are presented in the Table \ref{tab:perf_evaluation}.
\begin{table*}
\centering
\captionsetup{width=1.0\linewidth}
\caption{Performance evaluation of the proposed approach and comparison with well established sampling techniques (SMOTE, ADASYN, SWIM,  MOCAS (NN), SIMO and WSIMO ). \emph{GICaPS-O} is written when only oversampling is applied and \emph{GICaPS} is written when both undersampling and  oversampling are applied.}
\label{tab:perf_evaluation}
\footnotesize{\begin{tabular}{ |p{2cm}||p{2.0cm}|p{2.0cm}|p{2.0cm}|p{2.0cm}|p{2.0cm}|p{2.0cm}| }
 \hline
 \multicolumn{7}{|c|}{Evaluation matrices and comparison with different datasets} \\
 \hline
 Dataset & Methods & OA & Precision & Recall & F-measure & G-Mean\\
 \hline
   
\rowcolor{LightCyan}
 Abalone   & GICaPS-O  &\bf{96}&   \bf{96.80} & \bf{95.99} &\bf{96.32} & \bf{96.39} \\
 \cline{2-7}
  
      & SMOTE    &94.57&  89.3 &83.12 &75.71&86.15 \\
   \cline{2-7}
      & ADASYN    &   88.6&   32.21 & 48.16 &33.5 &39.38 \\
      & SWIM\cite{sharma2018synthetic} & -  &- & -&- & 72.3 \\
      
       & MOCAS (NN)\cite{lin2018minority} & -  &73.4 &44.7&54.6 & 82.7 \\
      
  \hline
\rowcolor{LightCyan} 
 Spambase   & GICaPS    & \bf{91.95}&    \bf{92.38} & \bf{92.38} & \bf{92.38} & \bf{92.38} \\
 \cline{2-7}
 
      & SMOTE    &89.95&  93.09 &90.14 & 91.46 &
91.60 \\
   \cline{2-7}
      & ADASYN    & 87.35&   88.81 &88.81 & 88.81 & 88.81\\
& SWIM\cite{sharma2018synthetic} & -  &- & -&- & 68.5 \\  
  \hline 
\rowcolor{LightCyan}  
  Glass & GICaPS-O    &\bf{96.38}&    97.02 & \bf{96.38} &  \bf{96.50} & \bf{96.7} \\
 \cline{2-7}

      & SMOTE    & 96.22&  \bf{97.08} & 96.22 & 96.46 & 96.65\\
   \cline{2-7}
      & ADASYN    & 91.51&   45.3 & 77.50 &43.32 & 59.25 \\
      & Cost sensitive & -  &- & -&- & 91.40 \\

	& SIMO \cite{piri2018synthetic}& -  &- & -&- & 92.84 \\
  
 	 & WSIMO\cite{piri2018synthetic}& -  &- & -&- & 92.86 \\
 & MOCAS (NN)\cite{lin2018minority} & -  &91.4 &82.8&86.2 & 92.3 \\ 	
 	 \hline 
\rowcolor{LightCyan}
Ionosphere & GICaPS-O    &\bf{88.50}&   90.38 & \bf{88.49}& \bf{89.00} & \bf{89.43} \\
 \cline{2-7}

      & SMOTE    & 84.65&   \bf{91.45} &84.63 &  87.06 & 87.97\\
   \cline{2-7}
      & ADASYN    & 81.60&  84.43 & 81.61 & 80.15 & 83.01\\
 & Cost sensitive & -  &- & -&- & 83.19 \\

	& SIMO \cite{piri2018synthetic} & -  &- & -&- & 84.69 \\
  
 	 & WSIMO \cite{piri2018synthetic}& -  &- & -&- & 84.99 \\  
  \hline 
  
\rowcolor{LightCyan}  
  Sonar & GICaPS-O    & \bf{85.55}&   \bf{90.39} &  \bf{90.48} &  \bf{90.28} & \bf{90.43} \\
 \cline{2-7}

      & SMOTE    &80.10&  87.39 &82.21 &84.37& 84.76\\
   \cline{2-7}
      & ADASYN    & 12.8&  11.23 & 12.79 & 11.76 & 11.98\\
  \hline 

\rowcolor{LightCyan}
  Wine & GICaPS-O    & \bf{89.67}&    \bf{90.41} & \bf{89.66} & \bf{89.84}  & \bf{90.03} \\
 \cline{2-7}
  & SMOTE    &83.60& 96.06 & 83.60 & 87.22 & 89.61\\
   \cline{2-7}
      & ADASYN    &   20.98& 18.21 & 20.99 & 19.09 & 19.55 \\
& SWIM\cite{sharma2018synthetic} & -  &- & -&- & 73.0 \\ 
  \hline 
\rowcolor{LightCyan}
Pima Indian diabetes & GICaPS-O    &\bf{84.15}&   \bf{83.77} & \bf{84.17} & \bf{83.88} & \bf{83.97}\\
 
 \cline{2-7}
  
      & SMOTE    & 76.40 &  81.16 &  76.53 & 78.15 & 78.81\\
   \cline{2-7}
      & ADASYN    & 75.45 & 75.58 & 75.74 & 75.65 & 75.66  \\
 & SWIM\cite{sharma2018synthetic} & -  &- & -&- & 50.9 \\ 
& MOCAS (NN)\cite{lin2018minority} & -  &73.4 &60.1&65.9 & 73.4 \\ 	 
  \hline 
%
\rowcolor{LightCyan}
  Shuttle & GICaPS    & \bf{99.39}&   \bf{99.30} &  \bf{99.37} &  \bf{99.33} & \bf{99.33}\\ 
 \cline{2-7}
  & SMOTE    & 96.59&   82.35 & 96.60 & 86.29 & 89.19\\
   \cline{2-7}
      & ADASYN    &56.27&   55.84 & 56.25 & 55.42 & 
56.04 \\
  \hline 
\rowcolor{LightCyan}
   Fertility & GICaPS-O    &  \bf{96.60}& \bf{96.56} & \bf{96.56} & \bf{96.56} & \bf{96.56} \\
 \cline{2-7}
 
      & SMOTE    &76.5&  81.26 &78.33 &78.65 & 79.78\\
   \cline{2-7}
      & ADASYN    &81.75& 82.02 & 83.84 & 80.89 & 82.92 \\
  \hline 
\rowcolor{LightCyan}  
  Pain & GICaPS  &\bf{98.80}& \bf{98.79} &\bf{92.61} & \bf{98.79} & \bf{95.65} \\
 \cline{2-7}
 
    & SMOTE    &90.24&  90.55 & 90.22 &  90.33 & 90.38\\
   \cline{2-7}
      & ADASYN    &13.2&   12.70 & 13.19 & 17.45 & 12.94 \\
  \hline 
 
\end{tabular}}
\end{table*}

\section{Conclusions}
\label{sec:concl}

Data imbalance poses serious challenges for classifier performance
particularly in cases where minority class is of importance. This
problem is addressed in this paper by proposing a data processing
framework called GICaPS that uses \emph{geometric information-based
sampling} and \emph{class-prioritized synthesis} for undersampling and
oversampling data in majority and minority classes respectively. The
proposed undersampling algorithm uses an angular constraint to remove
redundant information in a majority class while ensuring that the
valuable information is not lost. This is ensured by restricting the
removal of data points only from other orthants. On the other hand,
the proposed oversampling method populates the minority class by
generating data that respects class boundaries. This is achieved by
avoiding data generation in the \emph{no-man's land} between the
classes. Mathematical expressions are derived for these constraints
and concepts, thereby providing a theoretical basis for these
algorithms. Pseudocodes for these algorithms are provided for easy
implementation. The superiority of the proposed data sampling
algorithms is established through rigorous performance comparison
analysis with the current state-of-the-art methods on 10 different
real-world datasets exhibiting high data imbalance. The future scope
of this work would involve extending these concepts to hybrid
algorithms to further improve the classification performance on
imbalance datasets.  

  \bibliographystyle{IEEEtran} 
  \bibliography{paper_sk}

\end{document}